\documentclass{article}
\usepackage{microtype}
\usepackage{graphicx}
\usepackage{booktabs} 

\usepackage{hyperref}

\usepackage[accepted]{icml2019}

\usepackage[titletoc]{appendix}
\usepackage{lipsum}
\usepackage{tabularx,ragged2e}

\usepackage{titlesec}
\usepackage{bm}
\usepackage{subfig}
\usepackage{soul}

\usepackage[utf8]{inputenc} 
\usepackage[T1]{fontenc}    
\usepackage{url}            
\usepackage{amsfonts}       
\usepackage{nicefrac}       
\usepackage{microtype}      
\usepackage{cancel}
\usepackage{caption}
\usepackage[acronym]{glossaries}
\usepackage{color}

\usepackage{amsmath,amsfonts,amssymb}
\usepackage{amsthm}
\usepackage{graphicx} 
\usepackage{cancel}
\usepackage{verbatim}
\usepackage{url}
\usepackage{xcolor}
\usepackage{multirow}
\usepackage{caption}
\usepackage{algorithm}
\usepackage{algorithmic}

\newcommand{\ica}{\hspace{0.25cm}}

\usepackage{enumitem}

\usepackage{capt-of}

\usepackage{algorithm}
\usepackage{algorithmic}

\usepackage{tikz}
\usetikzlibrary{bayesnet}
\usepackage{booktabs,siunitx}
\usetikzlibrary{arrows}

\newtheorem{example}{Example}

\newtheorem{defn}{Definition}
\newtheorem{prop}{Proposition}

\graphicspath{{./figures/}}

\newglossaryentry{sde}{%
  name={sde},%
  description={stochastic differential equation},%
  first={stochastic differential equation (SDE)},%
  firstplural={stochastic differential equations (SDEs)},%
  text={SDE},%
  plural={SDEs}
}

\newcommand{\norm}{\mathcal{N}}

\newcommand{\avec}{\mathbf{a}}
\newcommand{\mvec}{\mathbf{m}}
\newcommand{\xvec}{\mathbf{x}}
\newcommand{\zvec}{\mathbf{z}}
\newcommand{\fvec}{\mathbf{f}}

\newcommand{\yvec}{\mathbf{y}}

\newcommand{\Xvec}{\mathbf{X}}

\newcommand{\Kcal}{\mathcal{K}}

\newcommand{\I}{\mathbf{I}}

\newcommand{\kff}{\mathbf{K}_\mathbf{ff}}

\allowdisplaybreaks[4]

\newcommand*{\mathcolor}{}
\def\mathcolor#1#{\mathcoloraux{#1}}
\newcommand*{\mathcoloraux}[3]{%
  \protect\leavevmode
  \begingroup
    \color#1{#2}#3%
  \endgroup
}

\makeatletter
\newcommand{\mypm}{\mathbin{\mathpalette\@mypm\relax}}
\newcommand{\@mypm}[2]{\ooalign{%
		\raisebox{.1\height}{$#1+$}\cr
		\smash{\raisebox{-.6\height}{$#1-$}}\cr}}
\makeatother

\usepackage{cleveref}

\DeclareMathOperator*{\argmin}{argmin}

\newcommand{\CM}[1]{{\color{black} #1}}
\newcommand{\YL}[1]{{\color{black} #1}}

\icmltitlerunning{Variational Implicit Processes}

\begin{document}

\twocolumn[
\icmltitle{Variational Implicit Processes}



\icmlsetsymbol{equal}{*}

\begin{icmlauthorlist}
\icmlauthor{Chao Ma}{cam}
\icmlauthor{Yingzhen Li}{msr}
\icmlauthor{Jos{\'e} Miguel Hern{\'a}ndez-Lobato}{cam,msr}
\end{icmlauthorlist}

\icmlaffiliation{cam}{Department of Engineering, University of Cambridge, Cambridge, UK}
\icmlaffiliation{msr}{Microsoft Research Cambridge, Cambridge, UK}

\icmlcorrespondingauthor{Chao Ma}{cm905@cam.ac.uk}
\icmlcorrespondingauthor{Jos{\'e} Miguel Hern{\'a}ndez-Lobato}{jmh233@cam.ac.uk}

\icmlkeywords{Machine Learning, ICML}

\vskip 0.3in
]



\printAffiliationsAndNotice{}  

\begin{abstract}
We introduce the implicit processes (IPs), a stochastic process that places implicitly defined multivariate distributions over any finite collections of random variables. IPs are therefore highly flexible implicit priors over \emph{functions}, with examples including data simulators, Bayesian neural networks and non-linear transformations of stochastic processes. A novel and efficient approximate inference algorithm for IPs, namely the variational implicit processes (VIPs), is derived using generalised wake-sleep updates. 
This method returns simple update equations and allows scalable hyper-parameter learning with stochastic optimization.
Experiments show that VIPs return better uncertainty estimates and lower errors over existing inference methods for challenging models such as Bayesian neural networks, and Gaussian processes. 
\end{abstract}


\section{Introduction}
\label{sec:intro}

Probabilistic models with \emph{implicit distributions} as core components have recently attracted enormous interest in both deep learning and the approximate Bayesian inference communities. In contrast to \emph{prescribed probabilistic models} \cite{diggle1984monte} that assign \emph{explicit} densities to possible outcomes of the model, implicit models \emph{implicitly assign} probability measures by the specification of the \emph{data generating process}. One of the most well known implicit distributions is the generator of generative adversarial nets (GANs) \cite{goodfellow2014generative, arjovsky2017wasserstein} that transforms isotropic noise into high dimensional data, using neural networks. In approximate inference context, implicit distributions have also been used as flexible approximate posterior distributions \cite{rezende2015variational, liu2016two, tran2017deep,li2017approximate}.

This paper explores the extension of implicit models to Bayesian modeling of \emph{random functions}. 
Similar to the construction of Gaussian processes (GPs), an \emph{implicit process} (IP) assigns implicit distributions over any finite collections of random variables. \YL{Therefore IPs can be much more flexible than GPs when complicated models like neural networks are used for the implicit distributions.}
With an IP as the prior, we can directly perform (variational) posterior inference \emph{over functions} in a 
non-parametric fashion. \CM{This is beneficial for better-calibrated uncertainty estimates like GPs \citep{bui2016deep}.} \YL{It also avoids typical issues of inference in parameter space}, \CM{ that is, symmetric modes in the posterior distribution of Bayesian neural network \emph{weights}.}\YL{The function-space inference for IPs is achieved by our proposed \emph{variational implicit process} (VIP) algorithm, which addresses the intractability issues of implicit distributions.}

Concretely, our contributions are threefold: 

\begin{itemize}[noitemsep,topsep=0pt,parsep=0pt,partopsep=0pt]
\item We formalize implicit stochastic process priors over \emph{functions}, and prove its well-definedness in both finite and infinite dimensional cases. By allowing the usage of IPs with rich structures as priors ( e.g., data simulators and Bayesian LSTMs), our approach provides a unified and powerful Bayesian inference framework for these important but challenging deep models. 

\item We derive a novel and efficient variational inference framework that gives a closed-form approximation to the IP posterior. It does not rely on e.g.~density ratio/gradient estimators in implicit variational inference literature which can be inaccurate in high dimensions. Our inference method is computationally cheap, and it allows scalable hyper-parameter learning in IPs.

\item We conduct extensive comparisons between IPs trained with the proposed inference method, and GPs/BNNs/Bayesian LSTMs trained with existing variational approaches. Our method consistently outperforms other methods, and achieves state-of-the-art results on a large scale Bayesian LSTM inference task. 

\end{itemize}


\section{Implicit Stochastic Processes} \label{sec:ip}

In this section, we generalize GPs to implicit stochastic processes. Readers are referred to appendix \ref{sec:app_gp_review} for a detailed introduction, but briefly speaking, a GP defines the distribution of a random function $f$ by placing a multivariate Gaussian distribution $\norm(\fvec;\mvec,\kff)$ over any finite collection of function values $\fvec = (f(\xvec_1), ..., f(\xvec_N))^\top$ evaluated at any given finite collection of input locations $\Xvec = \{ \xvec_n \}_{n=1}^N$. Here $(\mvec)_n = m(\xvec_n)$ and $(\kff)_{n,n'} = \Kcal(\xvec_n,\xvec_{n'})$, and following Kolmogorov consistency theorem \citep{ito1984introduction}, the mean and covariance functions $m(\cdot)$, $\Kcal(\cdot,\cdot)$ are shared across all such finite collections. An alternative parameterization of GPs defines the sampling process as $\fvec\sim\norm(\fvec;\mvec,\kff) \Leftrightarrow \zvec \sim \norm(\zvec;0,\mathbf{I})$, $\fvec  = \mathbf{B}\zvec+
\mvec$, with $\kff=\mathbf{B}\mathbf{B}^\top$ the Cholesky decomposition of the covariance matrix. 
Observing this, we propose a generalization of the generative process by replacing the linear transform of the latent variable $\zvec$ with a nonlinear one. This gives the following formal definition of implicit stochastic process.
\begin{defn} [noiseless implicit stochastic processes]
An implicit stochastic process (IP) is a collection of random variables $f(\cdot)$, such that any finite collection $\fvec = (f(\xvec_1), ..., f(\xvec_N))^\top$ has joint distribution implicitly defined by the following generative process: 
\begin{equation}
\zvec \sim p(\zvec), \ \ f(\xvec_n)  = g_{\theta}(\xvec_n,\zvec), \ \ \forall \ \xvec_n \in \Xvec.
\label{eq:implicit_process_def}
\end{equation}
A function distributed according to the above IP is denoted as $f(\cdot)\sim \mathcal{IP}(g_\theta(\cdot,\cdot),p_\zvec)$.
\label{df:IP}
\end{defn}

Note that $\zvec \sim p(\zvec)$ could be infinite dimensional (such as samples from a Gaussian Process). Definition \ref{df:IP} is validated by the following propositions. 
\begin{prop}[Finite dimension case]
Let $\mathbf{z}$ be a finite dimensional vector. Then there exists a unique stochastic process, such that any finite collection of random variables has distribution implicitly defined by (\ref{eq:implicit_process_def}).
\label{prop1}
\end{prop}
\begin{prop}[Infinite dimension case]
Let $z(\cdot) \sim \mathcal{SP}(0,C)$ be a centered continuous stochastic process on $\mathcal{L}^2(\mathbb{R}^d)$ with covariance function $C(\cdot,\cdot)$. Then the operator 
$g(\xvec, z) = O_k(z)(\xvec) := h(\int_{\mathbf{x}} \sum_{l=0}^{M}K_l(\xvec, \xvec^\prime) z(\xvec^\prime) d\xvec^\prime),\ 0<M< +\infty$ 
defines a stochastic process if $K_l \in \mathcal{L}^2(\mathbb{R}^d \times \mathbb{R}^d)$ , $h$ is a Borel measurable, bijective function in $\mathbb{R}$ and there exist $0 \leq A < +\infty$ such that $|h(x)| \leq A|x|$ for $\forall x \in \mathbb{R}$. 
\label{prop2}
\end{prop}


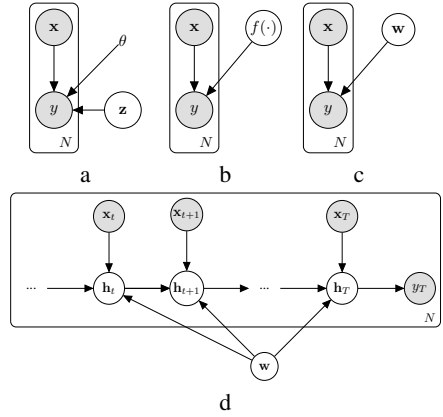
\begin{figure}
\centering
\subfloat[]{
\resizebox{1.6cm}{2cm}{
\begin{tikzpicture}


  \node[obs,minimum size=0.8cm]                               (y) {$y$};

  \node[const, above=of y, xshift=1.5cm] (theta) {$\mathbf{\theta}$};

  \node[obs, above=of y, xshift=0.0cm,minimum size=0.8cm]  (x) {$\mathbf{x}$};

  \node[latent, xshift=1.5cm,minimum size=0.8cm]            (z) {$\zvec$};


  \edge {x,theta,z} {y} ; %


  \plate {yx} {(x)(y)} {$N$} ;

\end{tikzpicture}
}
}
\hspace*{0.0cm}
\subfloat[]{
\resizebox{1.6cm}{2cm}{
\begin{tikzpicture}


  \node[obs,minimum size=0.8cm]                               (y) {$y$};

  \node[latent, above=of y, xshift=1.5cm] (w) {$f(\cdot)$};

  \node[obs, above=of y, xshift=0.0cm,minimum size=0.8cm]  (x) {$\mathbf{x}$};


  \edge {x,w} {y} ; %


  \plate {yx} {(x)(y)} {$N$} ;

\end{tikzpicture}
}
} 
\subfloat[]{
\resizebox{1.6cm}{2cm}{
\begin{tikzpicture}


  \node[obs,minimum size=0.8cm]                               (y) {$y$};

  \node[latent, above=of y, xshift=1.5cm] (w) {$\mathbf{w}$};

  \node[obs, above=of y, xshift=0.0cm,minimum size=0.8cm]  (x) {$\mathbf{x}$};


  \edge {x,w} {y} ; %


  \plate {yx} {(x)(y)} {$N$} ;

\end{tikzpicture}
}
}\\
\hspace*{0.0cm}
\subfloat[]{
\resizebox{5.8125cm}{2.5cm}{
\begin{tikzpicture}

  \node[const, xshift=0.0cm, minimum size=0.8cm] (o1) {$\text{...}$};
  \node[const, xshift=6.0cm, minimum size=0.8cm] (o2) {$\text{...}$};

  \node[latent, xshift=2.0cm, minimum size=0.8cm] (ht) {$\mathbf{h}_{t}$};
  \node[latent, xshift=4.0cm, minimum size=0.8cm] (ht1) {$\mathbf{h}_{t+1}$};
  \node[latent, xshift=8.0cm, minimum size=0.8cm] (hT) {$\mathbf{h}_{T}$};
  \node[latent, below=of o2, yshift=-0.2cm] (w) {$\mathbf{w}$};

  \node[obs, above=of ht, minimum size=0.8cm] (xt) {$\mathbf{x}_{t}$};
  \node[obs, above=of ht1, minimum size=0.8cm] (xt1) {$\mathbf{x}_{t+1}$};
  \node[obs, above=of hT, minimum size=0.8cm] (xT) {$\mathbf{x}_{T}$};
  \node[obs, xshift=10.0cm, minimum size=0.8cm] (y) {$y_{T}$};


  \edge {xt} {ht} ; %
  \edge {xt1} {ht1} ; %
  \edge {xT} {hT} ; %
  \edge {ht} {ht1} ; %
  \edge {o1} {ht}
  \edge {ht} {ht1}
  \edge {ht1} {o2}
  \edge {o2} {hT}
  \edge {hT} {y}
  \edge {w} {ht,ht1,hT}


  \plate {yx} {(xt)(xt1)(xT)(y)(ht)(ht1)(hT)(o1)} {$N$} ;

\end{tikzpicture}
}
}
\caption{Examples of IPs: (a) Neural samplers; (b) Warped GPs (c) Bayesian neural networks; (d) Bayesian RNNs. }
\label{IP_visual}
\end{figure}

Proposition \ref{prop1} is proved in appendix \ref{app:prop} using the Kolmogorov extension theorem. Proposition \ref{prop2} considers \emph{random functions} as the latent input $z(\cdot)$, and introduces a specific form of the transformation/operator $g$, so that the resulting collection of variables $f(\cdot)$ is still a valid stochastic process (see appendix \ref{app:prop2} for a proof).
Note this operator can be recursively applied to build highly non-linear operators over functions \citep{guss2016deep,williams1997computing,stinchcombe1999neural,le2007continuous,globerson2016learning}. 
These two propositions indicate that IPs form a rich class of priors over functions. Indeed, we visualize some examples of IPs in Figure \ref{IP_visual} with discussions as follows:
\begin{example}[Data simulators] 
Simulators, e.g.~physics engines and climate models, are omnipresent in science and engineering. These models encode laws of physics in $g_{\theta}(\cdot, \cdot)$, use $\zvec \sim p(\zvec)$ to explain the remaining randomness, and evaluate the function at input locations $\xvec$: $f(\xvec) = g_{\theta}(\xvec, \zvec)$. We define the \textbf{neural sampler} as a specific instance
of this class. In this case $g_{\theta}(\cdot, \cdot)$ is a neural network with weights $\theta$, i.e., $g_{\theta}(\cdot, \cdot) = \rm{NN}_{\theta}(\cdot, \cdot)$, and $p(\zvec) = \text{Uniform}([-a,a]^d)$.
\label{ex:ns}
\end{example}
\begin{example}[Warped Gaussian Processes] 
Warped Gaussian Processes \citep{snelson2004warped} is also an interesting example of IPs. Let $z(\cdot) \sim p(z)$ be a sample from a GP prior, and $g_{\theta}(\mathbf{x},z)$ is defined as $g_{\theta}(\mathbf{x},z) = h(z(\mathbf{x}))$, where $h(\cdot)$ is a one dimensional monotonic function. 
\end{example}
\begin{example}[Bayesian neural network] 
In a Bayesian neural network, the synaptic weights $W$ with prior $p(W)$ play the role of $\zvec$ in (\ref{eq:implicit_process_def}). A function is sampled by $W \sim p(W)$ and then setting $f(\xvec) = g_\theta(\xvec, W) = \rm{NN}_{W}(\xvec)$ for all $\xvec \in \Xvec$. In this case $\theta$ could be the hyper-parameters of the prior $p(W)$ to be tuned. 
\label{ex:bnn}
\end{example}
\begin{example}[Bayesian RNN] 
Similar to Example \ref{ex:bnn}, a Bayesian recurrent neural network (RNN) can be defined by considering its weights as random variables, and taking as function evaluation an output value generated by the RNN after processing the last symbol of an input sequence.
\label{ex:lstm}
\end{example}


\section{Variational Implicit Processes}


Consider the following regression model with an IP prior over the regression function:
\begin{equation}
\hspace{-0.02in}
f(\cdot) \sim \mathcal{IP}(g_\theta(\cdot,\cdot),p_\zvec), \ y = f(\xvec) + \epsilon, \ \epsilon \sim \norm(0,\sigma^2).
\label{eq:ip_regression_model}
\end{equation}
Equation (\ref{eq:ip_regression_model}) defines an implicit model $p(\mathbf{y}, \mathbf{f} | \xvec)$, which is intractable in most cases. Note that it is common to add Gaussian noise $\epsilon$ to an implicit model, e.g.~see the noise smoothing trick used in GANs \cite{sonderby2016amortised,salimans2016improved}.
Given an observed dataset $\mathcal{D} = \{\mathbf{X},\yvec\}$ and a set of test inputs $\mathbf{X}_*$, Bayesian predictive inference computes the predictive distribution $p(\yvec_*|\Xvec_*,\Xvec,\yvec,\theta)$, which itself requires interpolating over posterior $p(f|\Xvec, \yvec, \theta)$. Besides prediction, we also want to learn the model parameters $\theta$ and $\sigma$ by maximizing the marginal likelihood:
$
 \log p(\yvec | \mathbf{X},\theta) = \log \int_{\fvec} p(\yvec|\fvec)p(\fvec |\mathbf{X},\theta)d\fvec
$
, with $\fvec = f(\Xvec)$ being the evaluation of $f$ on the points in $\Xvec$. 
Unfortunately, both the prior $p(\fvec|\Xvec, \theta)$ and the posterior $p(f|\Xvec, \yvec, \theta)$ are intractable as the implicit process does not allow point-wise density evaluation, let alone the marginalization tasks. Therefore, to address these, we must resort to approximate inference.

We propose a generalization of the \emph{wake-sleep} algorithm \cite{hinton1995wake} to handle both intractabilities. This method returns (i) an approximate posterior distribution $q(f|\Xvec, \yvec)$ which is later used for predictive inference, and (ii) an approximation to the marginal likelihood $p(\yvec | \Xvec, \theta)$ for hyper-parameter optimization. We use the posterior of a GP to approximate the posterior of the IP, i.e.~$q(f | \Xvec, \yvec) = q_{\mathcal{GP}}(f|\Xvec, \yvec)$, since GP is one of the few existing tractable distributions over functions. A high-level summary of our algorithm is the following: 
\begin{itemize}
\item \textbf{\emph{Sleep phase}}:
sample function values $\fvec$ and noisy outputs $\yvec$ as indicated in (\ref{eq:ip_regression_model}). This \emph{dreamed} data is then used as the \emph{maximum-likelihood (ML)} target to fit a GP. This is equivalent to minimizing $\mathrm{D}_{\text{KL}}[p(\yvec, \fvec | \Xvec, \theta) || q_{\mathcal{GP}}(\yvec, \fvec | \Xvec)]$ for any possible $\Xvec$.
\item \textbf{\emph{Wake phase}}: 
The optimal GP posterior approximation  $q_{\mathcal{GP}}(\fvec|\Xvec, \yvec)$ obtained in the sleep phase is used to construct a variational approximation to $\log p(\yvec | \Xvec,\theta)$, which is then optimized
with respect to $\theta$.

\end{itemize}
Our approach has two key advantages. 
First, the algorithm has no explicit sleep phase computation, since the sleep phase optimization has an analytic solution that can be directly plugged into the wake-phase objective. 
Second, the proposed wake phase update is highly scalable, as it is equivalent to a Bayesian linear regression task with random features sampled from the implicit process.
With our wake-sleep algorithm, the evaluation of the implicit prior density is no longer an obstacle for approximate inference. We call this inference framework the \emph{variational implicit process} (VIP).
In the following sections we give specific details on both the wake and sleep phases.
\vspace{-0.6em}

\subsection{Sleep phase: GP posterior as variational distribution} \label{sec:amor}

This section proposes an approximation to the IP posterior $p(\fvec|\Xvec, \yvec, \theta)$. The naive variational inference \cite{jordan1999introduction} would require computing the joint distribution $p(\yvec, \fvec|\Xvec, \theta)$ which is intractable. However, sampling from this joint distribution is straightforward. We leverage this idea in the \emph{sleep phase} of our wake-sleep algorithm to approximate the joint distribution $p(\yvec, \fvec|\Xvec, \theta)$ instead.

Precisely, for any finite collection of variables $\fvec$ with their input locations $\Xvec$, we approximate $p(\yvec, \fvec|\Xvec, \theta)$ with a simpler distribution $q(\yvec, \fvec|\Xvec) = q(\yvec| \fvec) q(\fvec|\Xvec)$ instead. We choose $q(\fvec|\Xvec)$ to be a GP with mean and covariance functions $m(\cdot)$ and $\Kcal(\cdot, \cdot)$, respectively, and write the prior as $q(\fvec|\Xvec) = q_{\mathcal{GP}}(\fvec|\Xvec, m, \Kcal)$.
The sleep-phase update minimizes the following KL divergence: 
\begin{align}
&q^\star_{\mathcal{GP}} = \argmin_{m, \Kcal} \mathcal{U}(m,\Kcal), \\ &\text{with}\quad \mathcal{U}(m,\Kcal) = \mathrm{D}_{\text{KL}}[p(\yvec, \fvec | \Xvec, \theta) || q_{\mathcal{GP}}(\yvec, \fvec | \Xvec, m, \mathcal{K})]. \nonumber
\label{eq:sleep_objective}
\end{align}
We further assume $q(\yvec|\fvec) = p(\yvec|\fvec)$, which reduces $\mathcal{U}(m,\Kcal)$ to $\mathrm{D}_{KL}[p(\fvec | \Xvec, \theta) || q_{\mathcal{GP}}(\fvec | \Xvec, m, \mathcal{K})]$. In this case the optimal $m(\cdot)$ and $\mathcal{K}(\cdot, \cdot)$ are equal to the mean and covariance functions of the IP, respectively:
\begin{align}
m^{\star}(\xvec) &= \mathbb{E}[f(\xvec)], \\ \nonumber
\Kcal^{\star}(\xvec_1, \xvec_2) &= \mathbb{E}[ (f(\xvec_1) - m^{\star}(\xvec_1) ) (f(\xvec_2) - m^{\star}(\xvec_2) ) ].
\end{align}
Below we also write the optimal solution as $q^{\star}_{\mathcal{GP}}(\fvec | \Xvec, \theta) = q_{\mathcal{GP}}(\fvec | \Xvec, m^{\star}, \Kcal^{\star})$ to explicitly specify the \emph{dependency on prior parameters $\theta$} \footnote{This allows us to compute gradients w.r.t. $\theta$ through $m^{\star}$ and $\Kcal^{\star}$ using reparameterization trick (by definition of IP, $f(\xvec) = g_{\theta}(\xvec, \zvec)$), during the wake phase in Section \ref{sec:wake}.}.
In practice, the mean and covariance functions are estimated by by Monte Carlo, which leads to \emph{maximum likelihood} training (MLE) for the GP with \emph{dreamed} data from the IP. Assume $S$ functions are drawn from the IP: $f_s^\theta(\cdot) \sim \mathcal{IP}(g_\theta(\cdot,\cdot),p_\zvec), s=1,\ldots,S$. The optimum of $\mathcal{U}(m,\Kcal)$ is then estimated by the \emph{MLE solution}:
\begin{align}
m_{\text{MLE}}^\star(\xvec) &= \frac{1}{S} \sum_s f_s^\theta(\xvec), \label{eq:m} \\ 
\Kcal_{\text{MLE}}^\star(\xvec_1,\xvec_2)  &= \frac{1}{S}\sum_s \Delta_s(\xvec_1) \Delta_s(\xvec_2), \label{eq:K} \\ 
\nonumber \Delta_s(\xvec) &= f_s^\theta(\xvec)-m_{\text{MLE}}^\star(\xvec).
\end{align}
To reduce computational costs, the number of dreamed samples $S$ is often small. Therefore, we perform \emph{maximum a posteriori} instead of MLE, by putting an inverse Wishart process prior \cite{shah2014student} $\mathcal{IWP}(\nu,\Psi)$ over the GP covariance function $\mathcal{K}$ (Appendix \ref{app:wish}). 

The original sleep phase algorithm in \cite{hinton1995wake} also finds a posterior approximation by minimizing (\ref{eq:sleep_objective}). However, the original approach would define the $q$ distribution as $q(\yvec, \fvec|\Xvec) = p(\yvec|\Xvec, \theta) q_{\mathcal{GP}}(\fvec|\yvec, \Xvec)$, which builds a \emph{recognition model} that can be directly transfered for later inference. By contrast, we define $q(\yvec, \fvec|\Xvec) = p(\yvec|\fvec) q_{\mathcal{GP}}(\fvec|\Xvec)$, which corresponds to an approximation of the IP prior. In other words, we approximate an intractable generative model using another generative model with a GP prior and later, the resulting GP posterior $q^\star_{\mathcal{GP}}(\fvec|\Xvec, \yvec)$ is employed as the variational distribution. Importantly, we never explicitly perform the sleep phase updates, that is, the optimization of $\mathcal{U}(m, \mathcal{K})$, as there is an analytic solution readily available, which can potentially save a significant amount of computation. 

Another interesting observation is that the sleep phase's objective $\mathcal{U}(m,\Kcal)$ also provides an upper-bound to the KL divergence between the posterior distributions, 
$$\mathcal{J} = \mathrm{D}_{\text{KL}}[p(\fvec|\Xvec, \yvec, \theta) || q_{\mathcal{GP}}(\fvec|\Xvec, \yvec)].$$ One can show that $\mathcal{U}$ is an upper-bound of $\mathcal{J}$ according to the non-negativity and chain rule of the KL divergence:
\begin{equation}
\mathcal{U}(m,\Kcal) = \mathcal{J} + \mathrm{D}_{\text{KL}}[p(\yvec|\Xvec, \theta) || q_{\mathcal{GP}}(\yvec|\Xvec)] \geq \mathcal{J}.
\label{eq:sleep_phase_decomposition}
\end{equation}
Therefore, $\mathcal{J}$ is also decreased when the mean and covariance functions are optimized during the sleep phase. 
This bounding property justifies $\mathcal{U}(m,\Kcal)$ as a appropriate variational objective for posterior approximation. 


\subsection{Wake phase: a scalable approach to learning the model parameters $\theta$}\label{sec:wake}
In the wake phase of the original wake-sleep algorithm, the IP model parameters $\theta$ are optimized by maximizing a variational lower-bound on the log marginal likelihood $\log p(\yvec | \mathbf{X},\theta)$. Unfortunately, this requires evaluating the IP prior $p(\fvec |\Xvec, \theta)$ which is intractable. But recall from (\ref{eq:sleep_phase_decomposition}) that during the sleep phase $\mathrm{D}_{\text{KL}}[p(\yvec|\Xvec, \theta) || q_{\mathcal{GP}}(\yvec | \Xvec)]$ is also minimized. Therefore we directly approximate the log marginal likelihood using the \emph{optimal} GP from the sleep phase, i.e.
\begin{equation}
\log p(\yvec | \Xvec, \theta) \approx \log q^{\star}_{\mathcal{GP}}(\yvec | \Xvec, \theta).
\end{equation}
This again demonstrates the key advantage of the proposed sleep phase update via generative model matching.
Also it is a sensible objective for predictive inference as the GP returned by wake-sleep will be used for making predictions.

Similar to GP regression, optimizing $\log q^{\star}_{\mathcal{GP}}(\yvec | \Xvec, \theta)$ can be computationally expensive for large datasets. Therefore sparse GP approximation techniques \cite{snelson2006sparse,titsias2009variational,hensman2013gaussian,bui2016unifying} are applicable, but we leave them to future work and consider an alternative approach that is related to random feature approximations of GPs \cite{rahimi2008random,gal2015improving,gal2016dropout,balog2016mondrian,quia2010sparse}. 

Note that $\log q^\star_{\mathcal{GP}}(\yvec|\Xvec, \theta)$ can be approximated by the log marginal likelihood of a Bayesian linear regression model with $S$ randomly sampled dreamed functions, and a coefficient vector $\avec = (a_1,...,a_S)$:
\begin{equation}
\log q^\star_{\mathcal{GP}}(\yvec|\Xvec, \theta) \approx \log \int \prod_n q^\star (y_n|\xvec_n, \avec, \theta)p(\avec)d\avec, 
\label{eq:qgp}
\end{equation}
\vspace{-0.1in}
\begin{equation}
\centering
\begin{aligned}
q^\star (y_n|\xvec_n, \avec, \theta) &= \norm \left(y_n; \mu(\xvec_n, \avec, \theta),\sigma^2 \right), \\
\mu(\xvec_n, \avec, \theta) &= m^\star(\xvec_n) + \frac{1}{\sqrt{S}} \sum_s \Delta_s(\xvec_n) a_s, \\
\Delta_s(\xvec_n) = f_s^\theta(&\xvec_n) -m^\star(\xvec_n), \ p(\avec) = \norm(\avec; 0,\mathbf{I}).
\label{eq:bayesian_lr}
\end{aligned}
\end{equation}
For scalable inference, we follow \citet{li2017dropout} to approximate (\ref{eq:qgp}) by the $\alpha$-energy (see Appendix \ref{sec:alpha}), with $q_{\varphi}(\avec) = \mathcal{N}(\avec; \bm{\mu}, \bm{\Sigma})$ and mini-batch data $\{ \xvec_m, y_m\} \sim \mathcal{D}^M$:
\begin{equation}
\begin{aligned}
& \log q^\star_{\mathcal{GP}}(\yvec|\Xvec, \theta) \approx \mathcal{L}_{\mathcal{GP}}^{\alpha}(\theta, \varphi) \\ 
&= \frac{N}{\alpha M}\sum_m^M \log \mathbb{E}_{q_{\varphi}(\avec)} \left[ q^\star (y_m|\xvec_m, \avec, \theta)^\alpha \right] \\
&- \mathrm{D}_{\text{KL}}[q_{\varphi}(\avec)||p(\avec)].
\label{eq:qa}
\end{aligned}
\end{equation}
See Algorithm \ref{alg:VIP} for the full algorithm.
When $\alpha \rightarrow 0$ the $\alpha$-energy reduces to the variational lower-bound, and empirically the $\alpha$-energy \CM{returns better approximations} when $\alpha > 0$. For Bayesian linear regression (\ref{eq:bayesian_lr}) the exact posterior of $\avec$ is a multivariate Gaussian, which justifies our choice of $q_{\varphi}(\avec)$.
Stochastic optimization is applied to optimize $\theta$ and $\varphi$ jointly, \CM{making our method highly scalable.}

\begin{algorithm}[t]
\caption{Variational Implicit Processes (VIP)}
\label{alg:VIP}
 {\bfseries Require:} data $\mathcal{D} = (\Xvec, \yvec)$; IP $\mathcal{IP}(g_{\theta}(\cdot, \cdot), p_{\zvec})$; variational distribution $q_{\varphi}(\avec)$; hyper-parameter $\alpha$
\begin{algorithmic}[1]
 \WHILE{not converged}
 \STATE sample mini-batch $\{(\xvec_m, y_m)\}_{m=1}^M \sim
 \mathcal{D}^M$
 \STATE sample $S$ function values: \\$\zvec_s \sim p(\zvec), f_s^\theta(\xvec_m) = g_{\theta}(\xvec_m, \zvec_s) $
 \STATE solutions of \textbf{sleep phase}:\\ $m^\star(\xvec_m) = \frac{1}{S} \sum_{s=1}^S f_s^\theta(\xvec_m)$,\\ $\Delta_s(\xvec_m) = f_s^\theta(\xvec_m) - m^{\star}(\xvec_m)$
 \STATE compute the \textbf{wake phase} energy $\mathcal{L}_{\mathcal{GP}}^{\alpha}(\theta, \varphi)$ in (\ref{eq:qa}) using (\ref{eq:bayesian_lr})
 \STATE gradient descent on $\mathcal{L}_{\mathcal{GP}}^{\alpha}(\theta, \varphi)$ w.r.t~$\theta, \varphi$, via reparameterization tricks
 \ENDWHILE
\end{algorithmic}
\end{algorithm}

\subsection{Computational complexity and scalable predictive inference}
Assume the evaluation of a sampled function value $f(\xvec) = g_{\theta}(\xvec, \zvec)$ for a given input $\xvec$ takes $\mathcal{O}(C)$ time. The VIP has time complexity $\mathcal{O}(CMS + MS^2 +S^3)$ in training, where $M$ is the size of a mini-batch, and $S$ is the number of random functions sampled from  $\mathcal{IP}(g_\theta(\cdot,\cdot),p_\zvec)$. Note that approximate inference techniques in $\zvec$ space, e.g. mean-field Gaussian approximations to the posterior of Bayesian neural network weights \cite{blundell2015weight,hernandez2016black,li2017dropout}, also take $\mathcal{O}(CMS)$ time. Therefore when $C$ is large (typically the case for neural networks) the additional cost is often negligible, as $S$ is usually significantly smaller than the typical number of inducing points in sparse GP ($S=20$ in the experiments).

Predictive inference follows the standard GP equations to compute $q_{\mathcal{GP}}^\star(\fvec_*|\Xvec_*, \Xvec, \yvec, \theta^\star)$ on the test set $\Xvec_*$ with $K$ datapoints: $\fvec_* | \Xvec_*, \Xvec, \yvec \sim \mathcal{N}(\fvec_*; \mvec_*, \mathbf{\Sigma}_*)$,
\begin{equation}
\hspace{-0.05in}
\begin{aligned}
\mvec_* &= m^\star(\Xvec_*)+ \mathbf{K}_{* \fvec} ( \kff + \sigma^2\I )^{-1} ( \yvec-m^\star(\Xvec)), \\
\mathbf{\Sigma}_* &= \mathbf{K}_{**} - \mathbf{K}_{* \fvec}( \kff + \sigma^2\I )^{-1} \mathbf{K}_{\fvec *}.
\label{eq:gp_pred_mean_var}
\end{aligned}
\end{equation}
Recall that the optimal variational GP approximation has mean and covariance functions defined as (\ref{eq:m}) and (\ref{eq:K}), respectively, which means that $\kff$ has rank $S$. Therefore predictive inference requires both function evaluations and matrix inversion, which costs $\mathcal{O}(C(K+N)S + NS^2+S^3)$ time. This complexity can be further reduced: note that the computational cost is dominated by $(\kff + \sigma^2 \mathbf{I})^{-1}$. Denote the Cholesky decomposition of the kernel matrix $\kff = \mathbf{B} \mathbf{B}^\top$. It is straightforward to show that in the Bayesian linear regression problem (\ref{eq:bayesian_lr}) the exact posterior of $\avec$ is $q(\avec|\Xvec, \yvec) = \mathcal{N}(\avec; \bm{\mu}, \mathbf{\Sigma})$, with
$\bm{\mu} = \frac{1}{\sigma^2} \mathbf{\Sigma} \mathbf{B}^\top (\yvec - \mvec), \sigma^2 \mathbf{\Sigma}^{-1} = \mathbf{B}^\top \mathbf{B} + \sigma^2 \mathbf{I}.$
Therefore the parameters of the GP predictive distribution in (\ref{eq:gp_pred_mean_var}) are reduced to:
\begin{align}
\mathbf{m}_* & = m^\star(\Xvec_*) + \bm{\phi}_*^\top \bm{\mu}, \
\mathbf{\Sigma}_* = \bm{\phi}_*^\top \mathbf{\Sigma} \mathbf{\phi}_*,
\label{eq:scale}
\end{align}
with the elements in $\bm{\phi}_*$ as $(\bm{\phi}_*)_s = \Delta_s(\xvec_*) / \sqrt{S}$.
This reduces the prediction cost to $\mathcal{O}(CKS + S^3)$, which is on par with e.g.~conventional predictive inference techniques for Bayesian neural networks that also cost $\mathcal{O}(CKS)$. 
In practice we use the mean and covariance matrix from $q(\avec)$ to compute the predictive distribution.
Alternatively one can directly sample $\avec \sim q(\avec)$ and compute $\fvec_* = \sum_{s=1}^S a_s f_s^\theta(\Xvec_*)$, which is also an $\mathcal{O}(CKS + S^3)$ inference approach but would have higher variance.


\section{Experiments}
 In this section, we test the capability of VIPs with various tasks, including time series interpolation, Bayesian NN/LSTM inference, and Approximate Bayesian Computation (ABC) with simulators,etc. When the VIP is applied to Bayesian NN/LSTM (Example \ref{ex:bnn}-\ref{ex:lstm}), the prior parameters over each weight are tuned individually. We use $S=20$ for VIP unless noted otherwise. We focus on comparing VIPs as an \emph{inference method} to other Bayesian approaches, with detailed experimental settings presented in Appendix \ref{app:imp}. 

\subsection{Synthetic example}\label{sec:toy}
We first assess the behaviours of VIPs, including its quality of uncertainty estimation and the ability to discover structures under uncertainty. The synthetic training set is generated by first sampling 300 inputs $x$ from $\mathcal{N}(0, 1)$. Then, for each $x$ obtained, the corresponding target $y$ is simulated as $y = \frac{\cos5x}{|x|+1}+ \epsilon, \ \epsilon \sim \norm(0,0.1)$. The test set consists of $10^3$ evenly spaced points on $[-3,3]$.  We use an IP with a Bayesian neural network (1-10-10-1 architecture) as the prior. We use $\alpha=0$ for the wake-step training. We also compare VIP with the exact full GP with \emph{optimized} compositional kernel (RBF+Periodic), and another BNN with identical architecture but trained using variational dropout (VDO) with dropout rate $p= 0.99$ and length scale $l = 0.001$. The (hyper-)parameters are optimized using 500 epochs (batch training) with Adam optimizer (learning rate = 0.01).

\begin{figure*}[t]
\begin{minipage}[t]{0.67\textwidth}
\captionsetup[subfigure]{labelformat=empty}
\captionsetup[subfloat]{captionskip=-40pt}
\subfloat[]{\hspace{-0.22in}
\includegraphics[scale=0.29]{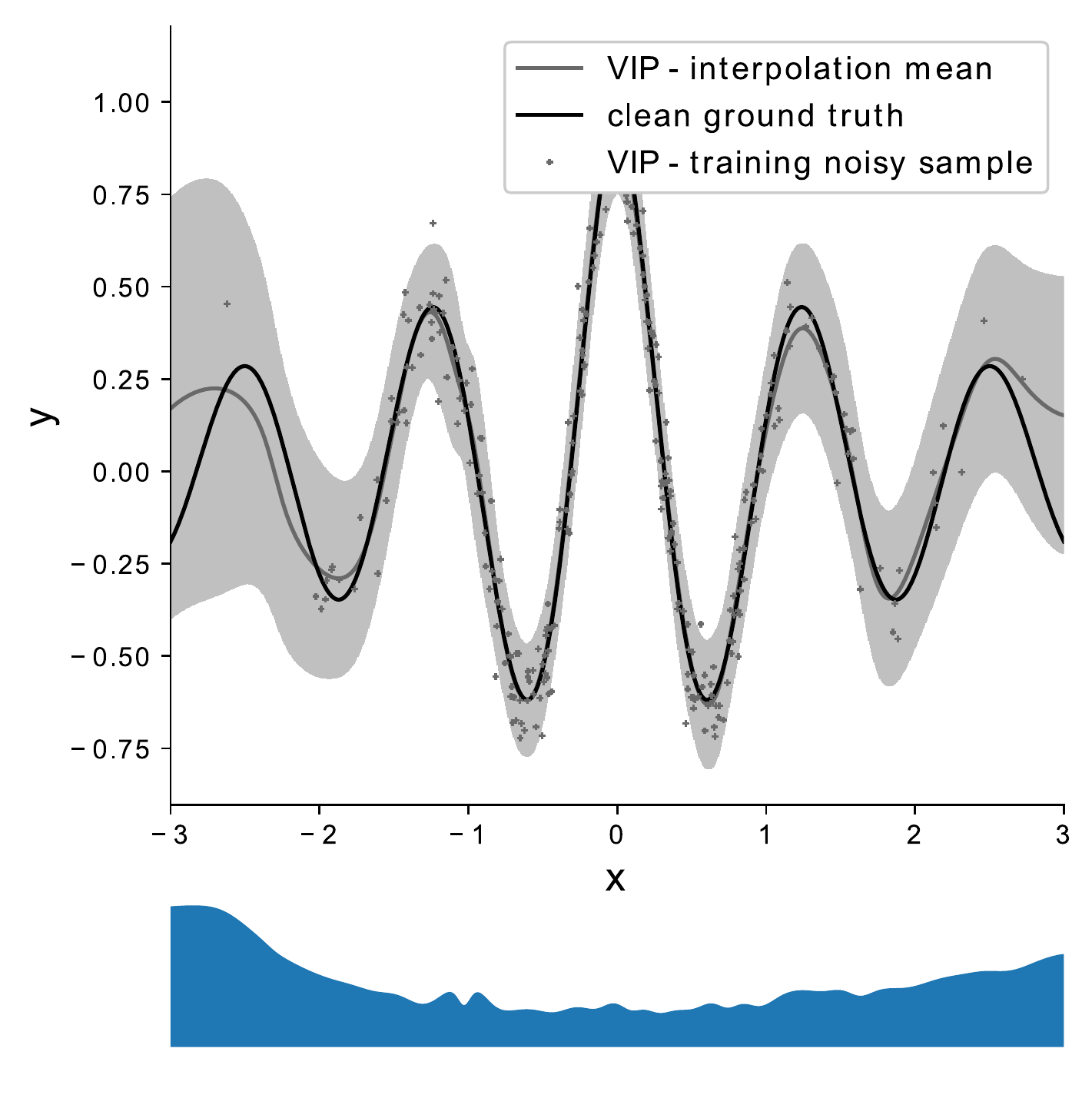}
\hspace{-11pt}
\includegraphics[scale=0.29]{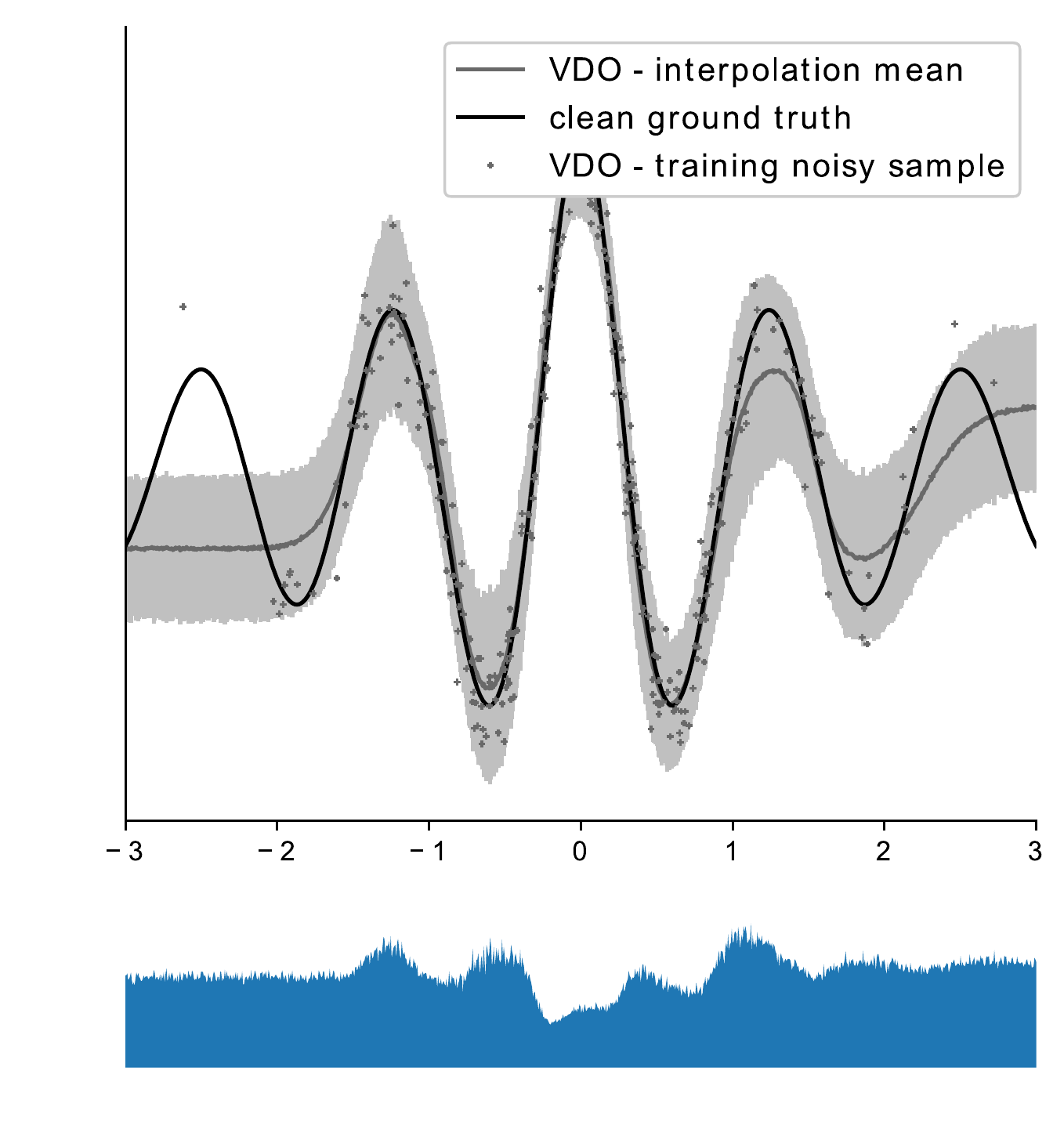}
\hspace{-11pt}
\includegraphics[scale=0.29]{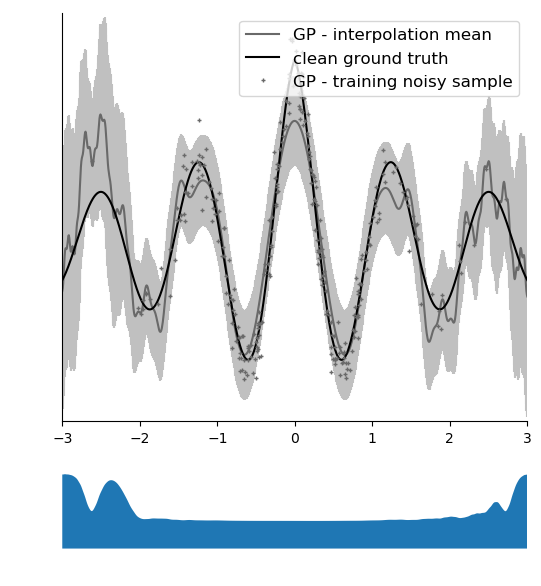}
\label{fig:toy_a}
}
\vspace*{-10pt}
\caption{First row: \small Predictions returned from VIP (\textbf{left}), VDO (\textbf{middle}) and exact GP with RBF + Periodic kernel (\textbf{right}), respectively. \textbf{Dark grey dots}: noisy observations; \textbf{dark line}: clean ground truth function; \textbf{dark gray line}: predictive means; \textbf{Gray shaded area}: confidence intervals with 2 standard deviations. \textbf{Second row}: Corresponding predictive uncertainties. }
\label{fig:toy1}
\end{minipage}
\hspace{3pt}
\begin{minipage}[t]{0.3\textwidth}
\vspace{-5pt}
\includegraphics[width=1.05\textwidth]{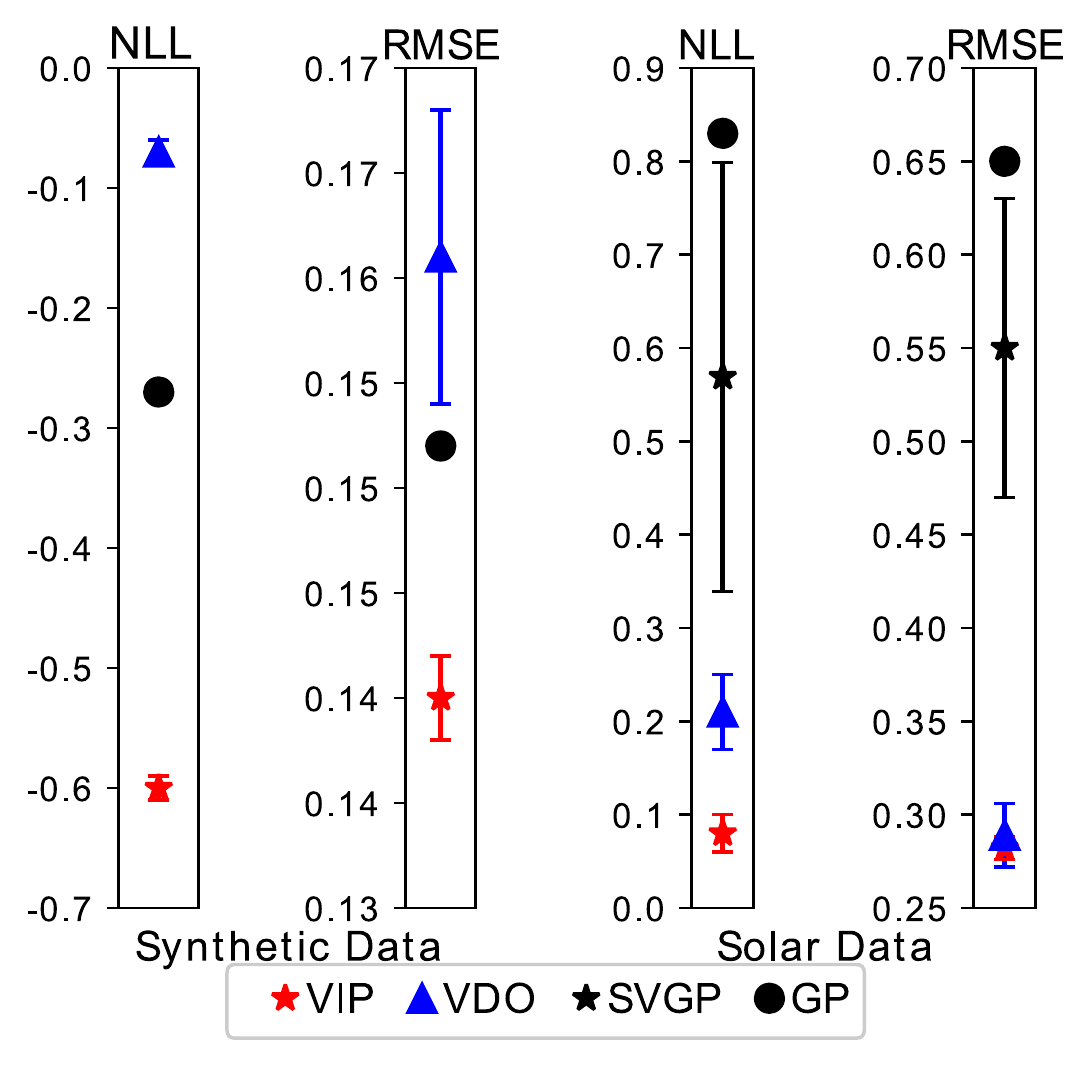}
\vspace*{-25pt}
\caption{Test performance on synthetic example (\textbf{left two}) and solar irradiance interpolation (\textbf{right two})}
\label{fig:toy2}
\end{minipage}
\vspace{-0.1in}
\end{figure*}

Figure \ref{fig:toy1} visualizes the results. Compared with VDO and the full GP, the VIP predictive mean recovers the ground truth function better. Moreover, VIP provides the best predictive uncertainty, especially when compared with VDO: it increases smoothly when $|x| \rightarrow 3$, where training data is sparse around there. \CM{Although the composition of periodic kernel helps the full GP to return a better predictive mean than VDO (but worse than VIP)}, it still over-fits to the data and returns a poor uncertainty estimate around $|x| \approx 2.5$.

Test Negative Log-likelihood (NLL) and RMSE results reveal similar conclusions (see the left two plots in Figure \ref{fig:toy2}), where VIP significantly outperforms VDO and GP.

\subsection{Solar irradiance interpolation under missingness}
Time series interpolation is an ideal task to evaluate the quality of uncertainty estimate. We compare the VIP ($\alpha=0$) with a variationally sparse GP (SVGP, 100 inducing points), an exact GP and VDO on the solar irradiance dataset \cite{lean1995reconstruction}. The dataset is constructed following \cite{gal2015improving}, where 5 segments of length 20 are removed for interpolation. All the inputs are then centered, and the targets are standardized. We use the same settings as in Section \ref{sec:toy}, except that we run Adam with learning rate = 0.001 for 5000 iterations. Note that GP/SVGP predictions are reproduced directly from \cite{gal2015improving}.

\begin{figure}[t]
\captionsetup[subfigure]{labelformat=empty}
\captionsetup[subfloat]{captionskip=-20pt}
\subfloat[]{
\hspace{-0.5in}
\includegraphics[scale=0.37]{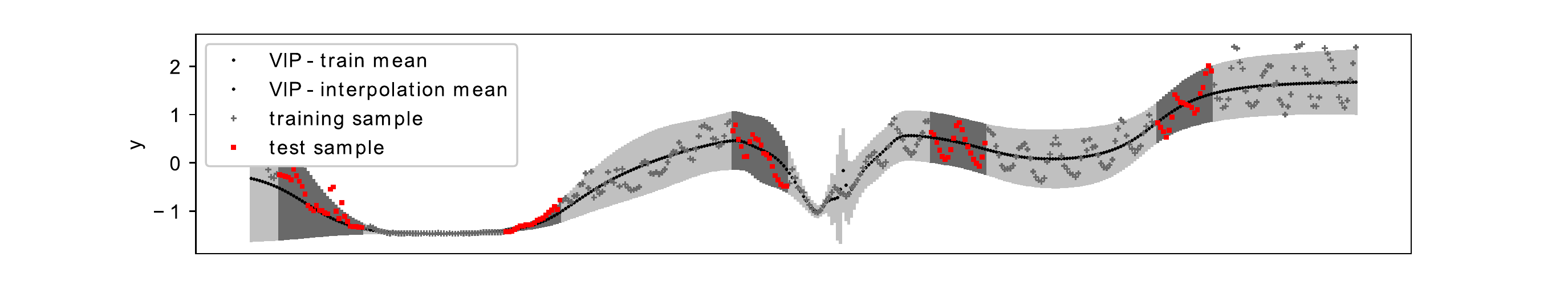}
}\\
\subfloat[]{
\hspace{-0.5in}
\includegraphics[scale=0.37]{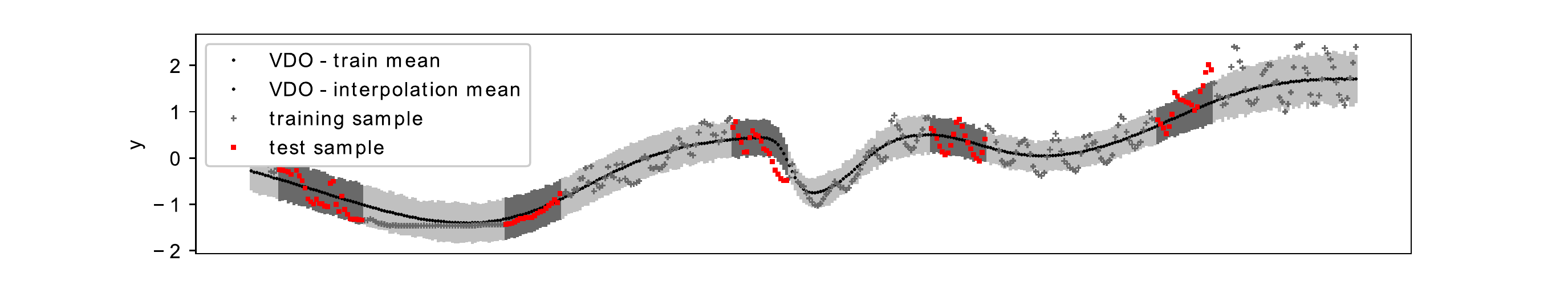}
}\\
\subfloat[]{
\hspace{-0.5in}
\includegraphics[scale=0.37]{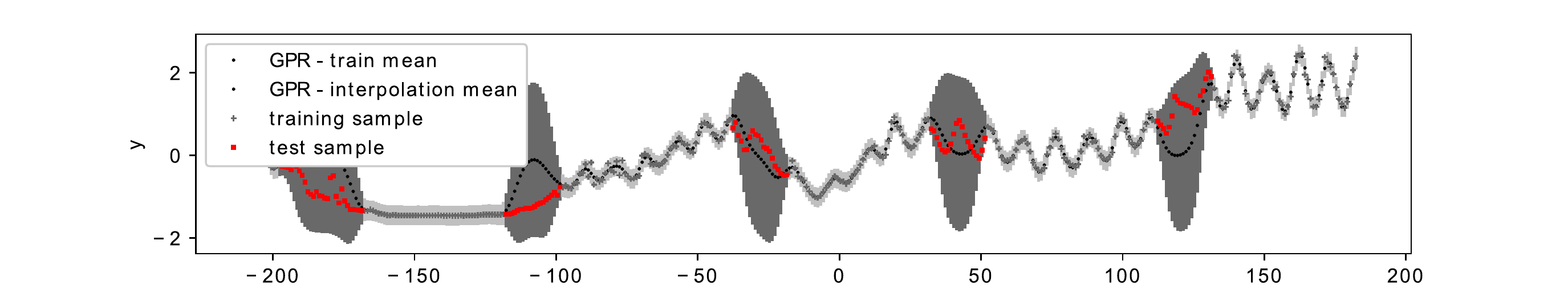}
}
\caption{Interpolations returned by VIP (\textbf{top}), variational dropout (\textbf{middle}), and exact GP (\textbf{bottom}), respectively. SVGP visualization is omitted as it looks nearly the same. Here \textbf{grey dots}: training data, \textbf{red dots}: test data, \textbf{dark dots}: predictive means, \textbf{light grey and dark grey areas}: Confidence intervals with 2 standard deviations of the training and test set, respectively. Note that our GP/SVGP predictions reproduces \cite{gal2015improving}. }
\label{fig:solar}
\end{figure}

Predictive interpolations are shown in Figure \ref{fig:solar}. We see that VIP and VDO give similar interpolation behaviors.  
However, VDO overall under-estimates uncertainty when compared with VIP, especially in the interval $[-100,200]$. VDO also incorrectly estimates the mean function around $x=-150$ where the ground truth there is a constant. On the contrary, VIP is able to recover the correct mean estimation around this interval with high confidence. GP methods recover the exact mean of the training data with high confidence, but they return poor estimates of predictive means for interpolation. 
Quantitatively, the right two plots in Figure \ref{fig:toy2} show that VIP achieves the best NLL/RMSE performance, again indicating that its returns high-quality uncertainties and accurate mean predictions.

\subsection{Predictive Performance: Multivariate regression}\label{exp:reg}
\YL{We apply the VIP inference to a Bayesian neural network (VIP-BNN, example \ref{ex:bnn}) and a neural sampler (VIP-NS, example \ref{ex:ns}) }, using real-world multivariate regression datasets from the UCI data repository \cite{lichman2013uci}. 
We mainly compare with the following BNNs baselines: variational Gaussian inference with reparameterization tricks \citep[VI,][]{blundell2015weight}, variational dropout \citep[VDO,][]{gal2016dropout}, and variational alpha dropout \citep{li2017dropout}. 
\CM{We also include the variational GP (SVGP, \cite{titsias2009variational}), exact GP and the functional BNNs (fBNN)\footnote{fBNN is a recent inference method designed for BNNs, where functional priors (GPs) are used to regularize BNN training. See related work for further discussions.}, and the results for fBNN is quoted from \citet{sun2018functional}}. 
All neural networks have two hidden layers of size 10, and are trained for 1,000 (except for fBNNs where the results cited use 2,000 epochs). The observational noise variance for VIP and VDO is tuned over a validation set, as detailed in Appendix \ref{app:imp}. The $\alpha$ value for both VIP and alpha-variational inference are fixed to 0.5, as suggested in \cite{hernandez2016black}. The experiments are repeated for 10 times on all datasets except \emph{Protein}, on which we report an averaged results across 5 repetitive runs. 
\begin{figure*}
\begin{minipage}[t]{0.75\textwidth}
\centering
\captionof{table}{Regression experiment: Average test negative log likelihood}
\vspace{-0.15in}
\label{tab:uci_llh}
\scalebox{0.7}{
\begin{tabular}{l@{\ica}r@{\ica}r@{\ica}r@{$\pm$}l@{\ica}r@{$\pm$}l@{\ica}r@{$\pm$}l@{\ica}r@{$\pm$}l@{\ica}r@{$\pm$}l@{\ica}r@{$\pm$}l@{\ica}|r@{$\pm$}l@{\ica}r@{$\pm$}l@{\ica}}
\hline
\bf{Dataset}&{N}&{D}&\multicolumn{2}{c}{\bf{VIP-BNN}}&\multicolumn{2}{c}{\bf{VIP-NS}}&\multicolumn{2}{c}{\bf{VI}}&\multicolumn{2}{c}{\bf{VDO}}&\multicolumn{2}{c}{\bf{$\alpha=0.5$}}&\multicolumn{2}{c|}{\bf{SVGP}}&\multicolumn{2}{c}{\bf{exact GP}} & \multicolumn{2}{c}{\bf{fBNN}}\\
\hline
boston&506&13&\textbf{2.45}&\textbf{0.04}&\textbf{2.45}&\textbf{0.03}&2.76&0.04&2.63&0.10&\textbf{2.45}&\textbf{0.02}&2.63&0.04&2.46&0.04 & 2.30 & 0.10\\
concrete&1030&8&\textbf{3.02}&\textbf{0.02}&3.13&0.02&3.28&0.01&3.23&0.01&3.06&0.03&3.4&0.01&3.05&0.02 & 3.09 & 0.01\\
energy&768&8&0.60&0.03&\textbf{0.59}&\textbf{0.04}&2.17&0.02&1.13&0.02&0.95&0.09&2.31&0.02&0.57&0.02 & 0.68 & 0.02\\
kin8nm&8192&8&\textbf{-1.12}&\textbf{0.01}&-1.05&0.00&-0.81&0.01&-0.83&0.01&-0.92&0.02&-0.76&0.00&\text{N/A}&0.00 &\text{N/A}&0.00\\
power&9568&4&2.92&0.00&2.90&0.00&2.83&0.01&2.88&0.00&\textbf{2.81}&\textbf{0.00}&2.82&0.00&\text{N/A}&0.00 &\text{N/A}&0.00 \\
protein&45730&9&\textbf{2.87}&\textbf{0.00}&2.96&0.02&3.00&0.00&2.99&0.00&2.90&0.00&3.01&0.00&\text{N/A}&0.00 &\text{N/A}&0.00\\
red wine&1588&11&\textbf{0.97}&\textbf{0.02}&1.20&0.04&1.01&0.02&\textbf{0.97}&\textbf{0.02}&1.01&0.02&0.98&0.02&0.26&0.03 & 1.04 & 0.01\\
yacht&308&6&\textbf{-0.02}&\textbf{0.07}&0.59&0.13&1.11&0.04&1.22&0.18&0.79&0.11&2.29&0.03&0.10&0.05 & 1.03 & 0.03\\
naval&11934&16&-5.62&0.04&-4.11&0.00&-2.80&0.00&-2.80&0.00&-2.97&0.14&\textbf{-7.81}&\textbf{0.00}&\text{N/A}&0.00 &\text{N/A}&0.00\\
\hline
\textbf{Avg.Rank}& & &\textbf{1.77}&\textbf{0.54}&2.77&0.57&4.66&0.28&3.88&0.38&2.55&0.37&4.44&0.66&\text{N/A}&0.00 &\text{N/A}&0.00\\
\hline
\end{tabular}
}

\vspace{0.1in}
\centering
\caption{Regression experiment: Average test RMSE}
\vspace{-0.15in}
\label{tab:uci_rmse}
\scalebox{0.73}{
\begin{tabular}{l@{\ica}r@{\ica}r@{\ica}r@{$\pm$}l@{\ica}r@{$\pm$}l@{\ica}r@{$\pm$}l@{\ica}r@{$\pm$}l@{\ica}r@{$\pm$}l@{\ica}r@{$\pm$}l@{\ica}|r@{$\pm$}l@{\ica}r@{$\pm$}l@{\ica}}
\hline
\bf{Dataset}&{N}&{D}&\multicolumn{2}{c}{\bf{VIP-BNN}}&\multicolumn{2}{c}{\bf{VIP-NS}}&\multicolumn{2}{c}{\bf{VI}}&\multicolumn{2}{c}{\bf{VDO}}&\multicolumn{2}{c}{\bf{$\alpha=0.5$}}&\multicolumn{2}{c|}{\bf{SVGP}}&\multicolumn{2}{c}{\bf{exact GP}}&\multicolumn{2}{c}{\bf{fBNN}}\\
\hline
boston&506&13&2.88&0.14&\textbf{2.78}&\textbf{0.12}&3.85&0.22&3.15&0.11&3.06&0.09&3.30&0.21&2.95&0.12 & 2.37 & 0.101\\
concrete&1030&8&\textbf{4.81}&\textbf{0.13}&5.54&0.09&6.51&0.10&6.11&0.10&5.18&0.16&7.25&0.15&5.31&0.15 & 4.93 & 0.18\\
energy&768&8&\textbf{0.45}&\textbf{0.01}&\textbf{0.45}&\textbf{0.05}&2.07&0.05&0.74&0.04&0.51&0.03&2.39&0.06&0.45&0.01 & 0.41 & 0.01\\
kin8nm&8192&8&\textbf{0.07}&\textbf{0.00}&0.08&0.00&0.10&0.00&0.10&0.00&0.09&0.00&0.11&0.01&\text{N/A}&0.00&\text{N/A}&0.00\\
power&9568&4&4.11&0.05&4.11&0.04&4.11&0.04&4.38&0.03&4.08&0.00&\textbf{4.06}&\textbf{0.04}&\text{N/A}&0.00&\text{N/A}&0.00\\
protein&45730&9&\textbf{4.25}&\textbf{0.07}&4.54&0.03&4.88&0.04&4.79&0.01&4.46&0.00&4.90&0.01&\text{N/A}&0.00&\text{N/A}&0.00\\
red wine&1588&11&\textbf{0.64}&\textbf{0.01}&0.66&0.01&0.66&0.01&\textbf{0.64}&\textbf{0.01}&0.69&0.01&0.65&0.01&0.62&0.01 & 0.67 & 0.01\\
yacht&308&6&\textbf{0.32}&\textbf{0.06}&0.54&0.09&0.79&0.05&1.03&0.06&0.49&0.04&2.25&0.13&0.35&0.04 & 0.60 & 0.06\\
naval&11934&16&\textbf{0.00}&\textbf{0.00}&\textbf{0.00}&\textbf{0.00}&0.38&0.00&0.01&0.00&0.01&0.00&\textbf{0.00}&\textbf{0.00}&\text{N/A}&0.00&\text{N/A}&0.00\\
\hline
\textbf{Avg.Rank}& & &\textbf{1.33}&\textbf{0.23}&2.22&0.36&4.66&0.33&4.00&0.44&3.11&0.42&4.44&0.72&\text{N/A}&0.00&\text{N/A}&0.00\\
\hline
\end{tabular}
}

\end{minipage}
\hspace{10pt}
\begin{minipage}[t]{0.2\textwidth}
\centering
\vspace{5pt}
\includegraphics[width=1.15\textwidth]{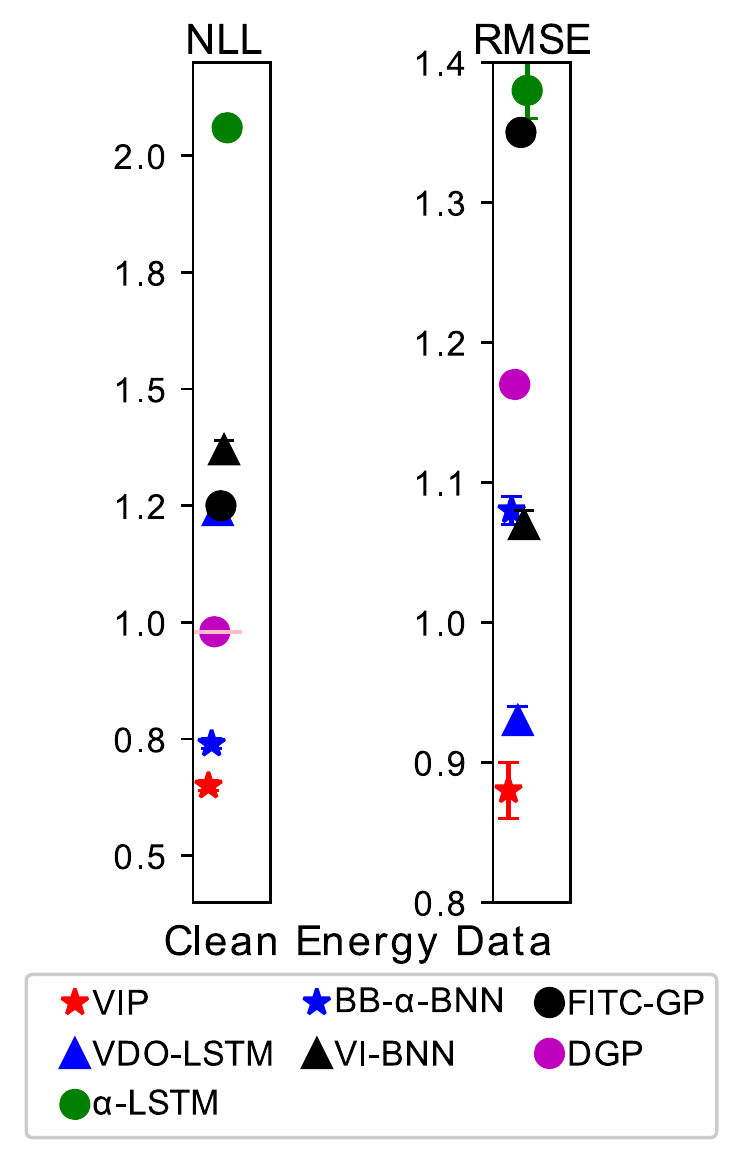}
\vspace*{-15pt}
\caption{\raggedleft Test performance on clean energy dataset}
\label{fig:cep}
\end{minipage}
\end{figure*}

Results are shown in Table \ref{tab:uci_llh} and \ref{tab:uci_rmse} with the best performances boldfaced. Note that our method is not directly comparable to exact (full) GP and fBNN in the last two columns. They are only trained on small datasets since they require the computation of the \emph{exact} GP likelihood, and fBNNs are trained for longer epochs. Therefore they are not included for the overall ranking shown in the last row of the tables. VIP methods consistently outperform other methods, obtaining the best test-NLL in 7 datasets, and the best test RMSE in 8 out of the 9 datasets. In addition, VIP-BNN obtains the best ranking among 6 methods. Note also that VIP marginally outperforms exact GPs and fBNNs (4 of 5 in NLLs), despite the comparison is not even fair. Finally, it is encouraging to see that, despite its general form, the VIP-NS achieves the second best average ranking in RMSE, outperforming many specifically designed BNN algorithms.

\subsection{Bayesian LSTM for predicting power conversion efficiency of organic photovoltaics molecules}\label{exp:lstm}
\CM{To demonstrate the scalability and flexibility} of VIP, we perform experiments with the Harvard Clean Energy Project Data, the world’s largest materials high-throughput virtual screening effort \cite{hachmann2014lead}. A large number of molecules of organic photovoltaics are scanned to find those with high power conversion efficiency (PCE) using quantum-chemical techniques. The target value of the dataset is the PCE of each molecule, and the input is the variable-length character sequence of the molecule structures. Previous studies have handcrafted \cite{pyzer2015learning, bui2016deep,hernandez2016black} or learned fingerprint features \cite{duvenaud2015convolutional} that transforms the raw string data into fixed-size features for prediction. 

We use a VIP with a prior defined by a Bayesian LSTM (200 hidden units) and $\alpha = 0.5$. We replicate the experimental settings in \citet{bui2016deep,hernandez2016black}, except that our method directly takes raw sequential molecule structure data as input. We compare our approach with a deep GP trained with expectation propagation \citep[DGP,][]{bui2016deep}, variational dropout for LSTM \citep[VDO-LSTM,][]{gal2016theoretically}, alpha-variational inference LSTM \citep[$\alpha$-LSTM,][]{li2017dropout}, BB-$\alpha$ on BNN \cite{hernandez2016black}, VI on BNN \cite{blundell2015weight}, and FITC GP \citep{snelson2006sparse}. Results for the latter 4 methods are quoted from \citet{hernandez2016black, bui2016deep}. Results in Figure \ref{fig:cep} show that VIP significantly outperforms other baselines and hits a state-of-the-art result in test likelihood and RMSE.

\subsection{ABC example: the Lotka–Volterra model}
Finally, we apply the VIP on an Approximate Bayesian Computation (ABC) example with the Lotka–Volterra (L-V) model that models the continuous dynamics of stochastic population of a predator-prey system. An L-V model consists of 4 parameters $\theta = \{\theta_1, \theta_2, \theta_3, \theta_4 \}$ that controls the rate of four possible random events in the model:
\begin{align*}
\dot{y}=\theta_1 xy-\theta_2 y,\quad \dot{x}=\theta_3 x-\theta_4 xy,
\end{align*}
where $x$ is the population of the predator, and $y$ is the population of the prey. Therefore the L-V model is an implicit model, which allows the simulation of data but not the evaluation of model density.
We follow the setup of \cite{papamakarios2016fast} to select the ground truth parameter of the L-V model, so that the model exhibit a oscillatory behavior which makes posterior inference difficult. Then the L-V model is simulated for 25 time units with a step size of 0.05, resulting in 500 training observations. The prediction task is to extrapolate the simulation to the $[25,30]$ time interval.

We consider (approximate) posterior inference using two types of approaches: regression-based methods (VIP-BNN, VDO-BNN and SVGP), and ABC methods (MCMC-ABC \cite{marjoram2003markov} and SMC-ABC \cite{beaumont2009adaptive,bonassi2015sequential}). 
ABC methods first perform posterior inference in the parameter space, then use the L-V simulator with posterior parameter samples for prediction. By contrast, regression-based methods treat this task as an ordinary regression problem, where VDO-BNN fits an approximate posterior to the NN weights, and VIP-BNN/SVGP perform predictive inference directly in function space. 
Results are shown in Table \ref{tab:ABC}, where VIP-BNN outperforms others by a large margin in both test NLL and RMSE. More importantly, VIP is the only regression-based method that outperforms ABC methods, demonstrating its flexibility in modeling implicit systems.

\begin{table}[t]
\centering
\caption{ABC with the Lotka–Volterra model}
\vspace{-0.15in}
\scalebox{0.8}{
\begin{tabular}{p{1.in}p{0.4in}p{0.4in}p{0.4in}p{0.4in}p{0.4in}} \toprule
\textbf{Method} &VIP-BNN & VDO-BNN & SVGP & MCMC-ABC & SMC-ABC \\ \midrule
Test NLL & \textbf{0.485} & $1.25$&$1.266$&$0.717$&$0.588$ \\ 
Test RMSE& \textbf{0.094} &0.80 & 0.950 & 0.307 &0.357 \\ \bottomrule
\end{tabular}
}
\label{tab:ABC}
\end{table}


\section{Related Research}

In the world of nonparametric models, Gaussian Processes \citep[GPs,][]{rasmussen2006Gaussian} provide accurate uncertainty estimates on unseen data, making them popular choices for Bayesian modelling in the past decades. Unfortunately, the $\mathcal{O}(N^3)$ time and $\mathcal{O}(N^2)$ space complexities make GPs impractical for large-scale datasets, therefore people often resort to approximations  \citep{quinonero2005unifying, snelson2006sparse, titsias2009variational, hensman2013gaussian, bui2016unifying, saatcci2012scalable}. Another intrinsic issue is the limited representational power of GPs with stationary kernels, limiting the applications of GP methods to high dimensional data \cite{bengio2005curse}.

In the world of parametric modeling, deep neural networks are extremely flexible function approximators that enable learning from very high-dimensional and structured data \citep{bengio2009learning, hinton2006fast,salakhutdinov2009deep, krizhevsky2012imagenet,simonyan2014very}. 
As people starts to apply deep learning techniques to critical applications such as health care, 
uncertainty quantification of neural networks has become increasingly important. Although decent progress has been made for Bayesian neural networks (BNNs) \citep{denker1991transforming,hinton1993keeping, barber1998ensemble, neal2012Bayesian,graves2011practical, blundell2015weight, hernandez2015probabilistic, li2017dropout}, uncertainty in deep learning still remains an open challenge. 

Research in the \emph{GP-BNN correspondance} has been extensively explored in order to improve the understandings of both worlds \citep{neal1996priors, neal2012Bayesian, williams1997computing, hazan2015steps,gal2016dropout, lee2017deep, matthews2018Gaussian}. Notably, in \citet{neal1996priors,gal2016dropout} a one-layer BNN with non-linearity $\sigma(\cdot)$ and mean-field Gaussian prior is approximately equivalent to a GP with kernel function 
\begin{align*}
\Kcal_{\text{VDO}}(\xvec_1,\xvec_2) = \mathbb{E}_{p(\mathbf{w})p(b)}[\sigma(\mathbf{w}^\top\xvec_1+b)\sigma(\mathbf{w}^\top\xvec_2+b)].
\end{align*}
Later \citet{lee2017deep} and \citet{matthews2018Gaussian} showed that a deep BNN is approximately equivalent to a GP with a \emph{compositional kernel} \cite{cho2009kernel, heinemann2016improper, daniely2016toward,poole2016exponential} that mimic the deep net. These approaches allow us to construct expressive kernels for GPs \cite{krauth2016autogp}, or conversely, exploit the \emph{exact} Bayesian inference on GPs to perform exact Bayesian prediction for BNNs \cite{lee2017deep}. The above kernel is compared with equation (\ref{eq:K}) in Appendix \ref{sec:app_compare}.

Alternative schemes have also been investigated to exploit deep structures for GP model design. These include: (1) \emph{deep GPs} \cite{damianou2013deep,bui2016deep}, where compositions of GP priors are proposed to represent prior over compositional functions; (2) 
the search and design of kernels for accurate and efficient learning \cite{van2017convolutional, duvenaud2013structure, tobar2015learning,beck2017learning,samo2015string},  and (3) \emph{deep kernel learning} that uses deep neural net features as the inputs to GPs \cite{hinton2008using,wilson2016stochastic, al2017learning, bradshaw2017adversarial, iwata2017improving}. Frustratingly, the first two approaches still struggle to model high-dimensional structured data such as texts and images; and the third approach is only Bayesian w.r.t.~the last output layer. 

The intention of our work is not to understand BNNs as GPs, nor to use deep learning to help GP design. Instead we directly treat a BNN as an instance of implicit processes (IPs), and the GP is used as a \emph{variational distribution} to assist predictive inference.
This approximation does not require previous assumptions in the GP-BNN correspondence literature \cite{lee2017deep,matthews2018Gaussian} nor the conditions in compositional kernel literature. 
Therefore the VIP approach also retains some of the benefits of Bayesian nonparametric approaches, and avoids issues of weight-space inference such as symmetric posterior modes.

\CM{To certain extent, the approach in \citet{flam2017mapping} resembles an inverse of VIP by encoding properties of GP priors into BNN weight priors, which is then used to regularize BNN inference. This idea is further investigated by a concurrent work on functional BNNs \cite{sun2018functional}, where GP priors are directly used to regularize BNN training through gradient estimators \cite{shi2018spectral}.}

Concurrent work of neural process \cite{garnelo2018neural} resembles the neural sampler, a special case of IPs. However, it performs inference in $\mathbf{z}$ space using the variational auto-encoder approach \cite{kingma2013auto,rezende2014stochastic}, which is not applicable to other IPs such as BNNs. By contrast, the proposed VIP approach applies to any IPs, and performs inference in function space. In the experiments we also show improved accuracies of the VIP approach on neural samplers over many existing Bayesian approaches.

\section{Conclusions}
We presented a variational approach for learning and Bayesian inference over function space based on implicit process priors. It provides a powerful framework that combines the rich flexibilities of implicit models with the well-calibrated uncertainty estimates from (parametric/nonparametric) Bayesian models. As an example, with BNNs as the implicit process prior, our approach outperformed many existing GP/BNN methods and achieved significantly improved results on molecule regression data.
Many directions remain to be explored.
Better posterior approximation methods beyond GP prior matching in function space will be designed.
Classification models with implicit process priors will be developed. Implicit process latent variable models will also be derived in a similar fashion as Gaussian process latent variable models. Future work will investigate novel inference methods for models equipped with other implicit process priors, e.g.~data simulators in astrophysics, ecology and climate science.

\subsection*{Acknowledgements} 
We thank Jiri Hron, Rich Turner, Andrew Foong, David Burt, Wenbo Gong and Cheng Zhang for discussions. Chao Ma thanks Microsoft Research donation, and Natural Science Foundation of Guangdong Province (2017A030313397) for supporting his research.

\nocite{langley00}

\bibliographystyle{icml2019}
\bibliography{REF}

\clearpage
\numberwithin{equation}{section}
 \begin{appendices}

\vspace{0.75cm}%
\begin{center}
{\huge Appendix}
\end{center}
\vspace{0.75cm}%

\section{Brief review of Gaussian processes}
\label{sec:app_gp_review}
Gaussian Processes \citep[GPs,][]{rasmussen2006Gaussian}, as a popular example of Bayesian nonparametrics, provides a principled probabilistic framework for non-parametric Bayesian inference over functions. This is achieved by imposing rich and flexible nonparametric priors over functions of interest. As flexible and interpretable function approximators, their Bayesian nature also enables GPs to provide valuable information of uncertainties regarding predictions for intelligence systems, all wrapped up in a single, \emph{exact} closed form solution of posterior inference.  

We briefly introduce GPs for regression. Assume that we have a set of observational data $\{ (\xvec_n,y_n \}_{n=1}^{N})$, where $\xvec_n$ is the $D$ dimensional input of $n$ th data point, and $y_n$ is the corresponding scalar target of the regression problem. A Gaussian Process model assumes that $y_n$ is generated according the following procedure: firstly a function $f(\cdot)$ is drawn from a Gaussian Process $\mathcal{GP}(m,k)$ (to be defined later). Then for each input data $\xvec_n$, the corresponding $y_n$ is then drawn according to:
\begin{align*}
y_n = f(\xvec_n) + \epsilon_n, \;\; \epsilon \sim \norm (0,\sigma^2), \;\;  {n=1,\cdots,N}
\end{align*}

A Gaussian Process is a nonparametric distribution defined over the space of functions, such that:
\begin{defn} [Gaussian Processes]
A Gaussian process (GP) is a collection of random variables, any finite number of
which have a joint Gaussian distributions. A Gaussian Process is fully specified by its mean function $m(\cdot): \mathbb{R}^D \mapsto \mathbb{R}$ and covariance function $\Kcal(\cdot,\cdot): (\mathbb{R}^D, \mathbb{R}^D) \mapsto \mathbb{R}$, such that any finite collection of function values $\fvec$ are distributed as Gaussian distribution $\norm(\fvec;\mvec,\kff)$, where $(\mvec)_n = m(\xvec_n)$, $(\kff)_{n,n'} = \Kcal(\xvec_n,\xvec_{n'})$.
\end{defn}
 
Now, given a set of observational data $\{ (\xvec_n,y_n) \}_{n=1}^{N}$, we are able to perform probabilistic inference and assign posterior probabilities over all plausible functions that might have generated the data. Under the setting of regression, given a new test point input data $\xvec_*$, we are interested in posterior distributions over $f_*$. Fortunately, this posterior distribution of interest admits a closed form solution $f_* \sim \norm(\mu_*,\Sigma_*)$:
\begin{align}
\mu_* = \mvec+K_{\xvec_* \fvec} ( \kff + \sigma^2\I )^{-1} ( \yvec-\mvec) 
\label{eq:GPm}
\end{align}
\begin{align}
\Sigma_* = K_{\xvec_* \xvec_*} - K_{\xvec_* \fvec}( \kff + \sigma^2\I )^{-1} K_{\fvec \xvec_*}
\label{eq:GPv}
\end{align}
In our notation, $(\yvec)_n = y_n$, $(K_{\xvec_* \fvec})_n = \Kcal(\xvec_*, \xvec_n)$, and $K_{\xvec_* \xvec_*} = \Kcal(\xvec_*, \xvec_*)$. Although the Gaussian Process regression framework is theoretically very elegant, in practice its computational burden is prohibitive for large datasets since the matrix inversion $( \kff + \sigma^2\I )^{-1}$ takes $\mathcal{O}(N^3)$ time due to Cholesky decomposition. Once matrix inversion is done, predictions in test time can be made in $\mathcal{O}(N)$ for posterior mean $\mu_*$ and $\mathcal{O}(N^2)$ for posterior uncertainty $\Sigma_*$, respectively. 

Despite the success and popularity of GPs (and other Bayesian non-parametric methods) in the past decades, their $\mathcal{O}(N^3)$ computation and $\mathcal{O}(N^2)$ storage complexities make it impractical to apply GPs to large-scale datasets. Therefore, people often resort to complicated approximate methods, e.g.~see \citet{seeger2003fast, quinonero2005unifying, snelson2006sparse, titsias2009variational, hensman2013gaussian, bui2016unifying, bui2014tree, saatcci2012scalable, cunningham2008fast, turner2010statistical}. 

Another critical issue to be addressed is the representational power of GP kernels. It has been argued that local kernels commonly used for nonlinear regressions are not able to obtain hierarchical representations for high dimensional data \cite{bengio2005curse}, which limits the usefulness of Bayesian non-parametric models for complicated tasks. A number of solutions were proposed, including deep GPs \citep{damianou2013deep,cutajar2016random,bui2016deep}, the design of expressive kernels \citep{van2017convolutional, duvenaud2013structure, tobar2015learning}, and the hybrid model with features from deep neural nets as the input of a GP \citep{hinton2008using,wilson2016stochastic}. However, the first two approaches still struggle to model complex high dimensional data such as texts and images; and in the third approach, the merits of fully Bayesian approach has been discarded.

\section{Brief review of variational inference, and the black-box $\alpha$-energy}\label{sec:alpha}
We give a brief review of modern variational techniques, including standard variational inference and black-box $\alpha$-divergence minimization (BB-$\alpha$), on which our methodology is heavily based. Considers the problem of finding the posterior distribution, $p(\theta|\mathcal{D},\tau)$, $\mathcal{D} = \{\xvec_n\}_{n=1}^{N}$ under the model likelihood $p(\xvec|\theta, \tau)$ and a prior distribution $p_0(\theta) $: 
\begin{align*}
p(\theta|\mathcal{D},\tau) \propto \frac{1}{Z} p_0(\theta) \prod_{n} p(\xvec_n|\theta, \tau).
\end{align*}
Here $\tau$ is the hyper-parameter of the model, which will be optimized by (approximate) maximum likelihood.

Variational inference \citep[VI,][]{jordan1999introduction} converts the above inference problem into an optimization problem, by first proposing a class of approximate posterior $q(\theta)$, and then minimize the KL-divergence from the approximate posterior to the true posterior $\mathcal{D}_{\text{KL}}[q||p]$. Equivalently, VI optimizes the following variational free energy, 
\begin{equation*}
\begin{aligned}
\mathcal{F}_{\text{VFE}} &= \log p(\mathcal{D}|\tau) - \mathcal{D}_{\text{KL}}[q(\theta)||p(\theta|\mathcal{D}, \tau)] \\ &= \mathbb{E}_{q(\theta)} \left[ \log \frac{p(\mathcal{D},\theta|\tau)}{q(\theta)} \right].
\end{aligned}
\end{equation*}

Built upon the idea of VI, BB-$\alpha$ is a modern black-box variational inference framework that unifies and interpolates between VI \cite{jordan1999introduction} and expectation propagation (EP)-like algorithms \cite{minka2001expectation, li2015stochastic}. BB-$\alpha$ performs approximate inference by minimizing the following $\alpha$-divergence \cite{zhu1995information} $\mathrm{D}_{\alpha}[p||q]$:
\begin{equation*}
\label{eq:alpha_divergence}
\mathrm{D}_{\alpha}[p||q] = \frac{1}{\alpha (1 - \alpha)} \left(1 - \int p(\theta)^{\alpha} q(\theta)^{1 - \alpha} d\theta \right).
\end{equation*}
$\alpha$-divergence is a generic class of divergences that includes the inclusive KL-divergence ($\alpha$=1, corresponds to EP), Hellinger distance ($\alpha$=0.5), and the exclusive KL-divergence ($\alpha$ = 0, corresponds to VI) as special cases. 

Traditionally, power EP \cite{minka2004power} optimizes an $\alpha$-divergence locally with exponential family approximation $q(\theta) \propto \frac{1}{Z} p_0(\theta) \prod_{n} \tilde{f}_n(\theta)$,$\tilde{f}_n(\theta) \propto \exp \left[ \lambda_n^{T}\phi(\theta) \right]$ via message passing. It converges to a fixed point of the so called \emph{power EP energy}:
\begin{align*}
&\mathcal{L}_{\text{PEP}}(\lambda_0,\{ \lambda_n \}) =  \log Z(\lambda_0) + (\frac{N}{\alpha}-1) \log Z(\lambda_q) \\
  &- \frac{1}{\alpha} \sum_{n=1}^{N} \log \int p(\xvec_n|\theta, \tau)^{\alpha} \exp\left[ (\lambda_q - \alpha \lambda_n)^T \phi(\theta) \right] d\theta,
\label{eq:energy}
\end{align*}
where $\lambda_q = \lambda_0 + \sum_{n=1}^N \lambda_n$ is the natural parameter of $q(\theta)$. On the contrary,  BB-$\alpha$ directly optimizes $\mathcal{L}_{\text{PEP}}$ with tied factors $\tilde{f}_n = \tilde{f}$ to avoid prohibitive local factor updates and storage on the whole dataset. This means $\lambda_n = \lambda$ for all $n$ and $\lambda_q = \lambda_0 + N \lambda$. Therefore instead of parameterizing each factors, one can directly parameterize $q(\theta)$ and replace all the local factors in the power-EP energy function by $\tilde{f}(\theta) \propto (q(\theta)/p_0(\theta))^{1/N}$.
After re-arranging terms, this gives the BB-$\alpha$ energy:
\begin{equation*}
\mathcal{L}_{\alpha}(q) = -\frac{1}{\alpha} \sum_n \log \mathbb{E}_q \left[ \left(  \frac{f_n(\theta) p_0(\theta)^{\frac{1}{N}}}{q(\theta)^{\frac{1}{N}}} \right)^{\alpha} \right].
\label{eq:bbalpha_original}
\end{equation*}
which can be further approximated by the following if the dataset is large \cite{li2017dropout}:
\begin{equation*}
\mathcal{L}_{\alpha}(q) = \mathcal{D}_{\text{KL}}[q||p_0] -\frac{1}{\alpha} \sum_n \log \mathbb{E}_q \left[ p(\xvec_n|\theta, \tau)^{\alpha} \right].
\label{eq:bbalpha_dropout}
\end{equation*}
The optimization of $\mathcal{L}_{\alpha}(q)$ could be performed in a black-box manner with reparameterization trick \cite{kingma2013auto}, Monte Carlo (MC) approximation and mini-batch training. Empirically, it has been shown that BB-$\alpha$ with $\alpha\neq 0$ can return significantly better uncertainty estimation than VI, and has been applied successfully in different scenarios \cite{li2017dropout, depeweg2016learning}.  From hyper-parameter learning (i.e., $\tau$ in $p(\xvec_n|\theta, \tau)$), it is shown in \citet{li2016renyi} that the BB-$\alpha$ energy $\mathcal{L}_{\alpha}(q)$ constitutes a better estimation of log marginal likelihood, $\log p(\mathcal{D}|\tau)$ when compared with the variational free energy. Therefore, for both inference and learning, BB-$\alpha$ energy is extensively used in this paper.



\section{Derivations}
\subsection{Proof of Proposition 1 (finite dimensional case)}\label{app:prop}
\textbf{Proposition 1.} \emph{If $\zvec$ is a finite dimensional random variable, then there exists a unique stochastic process, with finite marginals that are distributed exactly according to Definition \ref{df:IP}.}

\paragraph{Proof} 
Generally, consider the following noisy IP model:
\begin{align*}
f(\cdot) \sim \mathcal{IP}(g_\theta(\cdot,\cdot),p_\zvec), \ \ y_n = f(\xvec_n) + \epsilon_n, \ \epsilon_n \sim \norm(0,\sigma^2).
\end{align*}
For any finite collection of random variables $y_{1:n} = \{y_1, ..., y_n \}$, $\forall n$ we denote the induced distribution as $p_{1:n}(y_{1:n})$. 
Note that $p_{1:n}(y_{1:n})$ can be represented as $\mathbb{E}_{p(\zvec)}[ \prod_{i=1}^n \mathcal{N}(y_i; g(\xvec_i; \zvec), \sigma^2) ]$. 
Therefore for any $m < n$, we have
\begin{align*}
&\int p_{1:n}(y_{1:n}) d y_{m+1:n} \\
&= \int\int \prod_{i=1}^n \mathcal{N}(y_i; g(\xvec_i, \zvec), \sigma^2) p(\zvec) d\zvec d y_{m+1:n} \\
&= \int\int \prod_{i=1}^n \mathcal{N}(y_i; g(\xvec_i, \zvec), \sigma^2) p(\zvec) d y_{m+1:n} d\zvec \\
&= \int \prod_{i=1}^m \mathcal{N}(y_i; g(\xvec_i, \zvec), \sigma^2) p(\zvec) d\zvec = p_{1:m}(y_{1:m}).
\end{align*}
Note that the swap of the order of integration relies on that the integral is finite, which is true when the prior $p(\zvec)$ is proper.
Therefore, the marginal consistency condition of Kolmogorov extension theorem is satisfied. Similarly, the permutation consistency condition of Kolmogorov extension theorem can be proved as follows: assume $\pi(1:n) = \{\pi(1), ..., \pi(n) \}$ is a permutation of the indices $1:n$, then 
\begin{align*}
&p_{\pi(1:n)}(y_{\pi(1:n)}) \\
&= \int \prod_{i=1}^n \mathcal{N}(y_{\pi(i)}; g(\xvec_{\pi(i)}, \zvec), \sigma^2) p(\zvec) d\zvec \\
&= \int \prod_{i=1}^n \mathcal{N}(y_i; g(\xvec_i, \zvec), \sigma^2) p(\zvec) d\zvec = p_{1:n}(y_{1:n}).
\end{align*}
Therefore, by Kolmogorov extension theorem, there exists a unique stochastic process, with finite marginals that are distributed exactly according to Definition \ref{df:IP}. 

\qed

\subsection{Proof of Proposition 2 (infinite dimensional case)}\label{app:prop2}
\textbf{Proposition 2.} \emph{Let $z(\cdot) \sim \mathcal{SP}(0,C)$ be a centered continuous stochastic process on $ \mathcal{L}^2(\mathbb{R}^d)$ with covariance function $C(\cdot,\cdot)$. Then the operator 
$g(\mathbf{x}, z) = O(z)(\xvec) := h(\int \sum_{l=0}^{M}K_l(\xvec, \xvec^\prime) z(\xvec^\prime) d\xvec^\prime),\ 0< M < + \infty$ 
defines a stochastic process if $K_l \in \mathcal{L}^2(\mathbb{R}^d \times \mathbb{R}^d)$ , $h$ is a Borel measurable, bijective function in $\mathbb{R}$ and there exist $0 \leq A < + \infty$ such that $|h(x)| \leq A|x|$ for $\forall x \in \mathbb{R}$.  }

\paragraph{Proof} 
Since $\mathcal{L}^2(\mathbb{R}^d)$ is closed under finite summation, without loss of generality, we consider the case of $M=1$ where $O(z)(\xvec) = h(\int K(\xvec, \xvec^\prime) z(\xvec^\prime) d\xvec^\prime)$. According to Karhunen-Loeve expansion (K-L expansion) theorem \cite{loeve}, the stochastic process $z$ can be expanded as the \emph{stochastic} infinite series,
$$z(\xvec) = \sum_i^\infty Z_i \phi_i(\mathbf{x}), \quad \sum_i^\infty \lambda_i < +\infty. $$
Where $Z_i$ are zero-mean, uncorrelated random variables with variance $\lambda_i$. Here $\{ \phi_i \}_{i=1}^{\infty}$ is an orthonormal basis of $\mathcal{L}^2(\mathbb{R}^d)$ that are also eigen functions of the operator $O_C(z)$ defined by $O_C(z)(\xvec) = \int C(\mathbf{x}, \mathbf{x}^\prime) z(\mathbf{x}^\prime) d\mathbf{x}^\prime$. The variance $\lambda_i$ of $Z_i$ is the corresponding eigen value of $\phi_i(\mathbf{x})$.  

Apply the linear operator $$O_K(z)(\xvec) = \int K(\mathbf{x}, \mathbf{x}^\prime) z(\mathbf{x}^\prime) d\mathbf{x}^\prime$$ on this K-L expansion of $z$, we have:
\begin{align}
\nonumber O_K(z)(\xvec) & = \int K(\mathbf{x}, \mathbf{x}^\prime) z(\mathbf{x}^\prime) d\mathbf{x}^\prime \\ 
\nonumber & = \int K(\mathbf{x}, \mathbf{x}^\prime) \sum_i^\infty Z_i \phi_i(\mathbf{x}^\prime) d\mathbf{x}^\prime \\ 
\nonumber & =\sum_i^\infty Z_i \int K(\mathbf{x}, \mathbf{x}^\prime)   \phi_i(\mathbf{x}^\prime) d\mathbf{x}^\prime, \\
\label{eq:linear}
\end{align}
where the exchange of summation and integral is guaranteed by Fubini's theorem. Therefore, the functions $\{ \int_{\mathbf{x}} K(\mathbf{x}, \mathbf{x}^\prime)   \phi_i(\mathbf{x}^\prime) d\mathbf{x}^\prime \}_{i=1}^{\infty}$ forms a new basis of $\mathcal{L}^2(\mathbb{R}^d)$. To show that the stochastic series \ref{eq:linear} converge:
\begin{align*}
&||\sum_i^\infty Z_i \int K(\mathbf{x}, \mathbf{x}^\prime)   \phi_i(\mathbf{x}^\prime) d\mathbf{x}||^2_{\mathcal{L}^2} \\
& \leq ||O_K||^2||\sum_i^\infty Z_i \phi_i(\mathbf{x}^\prime)||^2_{\mathcal{L}^2} \\
& = ||O_K||^2 \sum_i^\infty ||Z_i||^2_2 ,
\end{align*}
where the operator norm is defined by 
$$||O_K|| := \inf \{c\geq 0: ||O_k(f)||_{\mathcal{L}^2} \leq c||f||_{\mathcal{L}^2},\ \forall f \in \mathcal{L}^2 (\mathbb{R}^d) \}.$$ 
This is a well defined norm since $O_K$ is a bounded operator ($K \in \mathcal{L}^2(\mathbb{R}^d \times \mathbb{R}^d)$). The last equality follows from the orthonormality of $\{\phi_i \}$. The condition $\sum_i^\infty \lambda_i < \infty$ further guarantees that $\sum_i^\infty ||Z_i||^2$ converges almost surely. Therefore, the random series (\ref{eq:linear}) converges in $\mathcal{L}^2 (\mathbb{R}^d)$ a.s.. 
 

Finally we consider the nonlinear mapping $h(\cdot)$. With $h(\cdot)$ a Borel measurable function satisfying the condition that there exist $0 \leq A < +\infty$ such that $|h(x)| \leq A|x|$ for $\forall x \in \mathbb{R}$, it follows that $h \circ O_K(z) \in \mathcal{L}^2 (\mathbb{R}^d)$.   In summary, $g = O_k(z) = h \circ O_K(z)$ defines a well-defined stochastic process on $\mathcal{L}^2 (\mathbb{R}^d)$. 

\qed

Despite of its simple form, the operator $g = h \circ O_K(z)$ is in fact the building blocks for many flexible transformations over functions \citep{guss2016deep,williams1997computing,stinchcombe1999neural,le2007continuous,globerson2016learning} . Recently \citet{guss2016deep} proposed the so called Deep Function Machines (DFMs) that possess universal approximation ability to nonlinear operators:

\begin{defn} [Deep Function Machines \cite{guss2016deep}]
A deep function machine $g = O_{DFM}(z,S)$ is a computational skeleton
S indexed by I with the following properties:
\begin{itemize}
\item Every vertex in $S$ is a Hilbert space $\mathbb{H}_l$ where $l \in I$.
\item If nodes $l\in A \subset I$ feed into $l^\prime$ then the activation on $l^\prime$ is denoted $y^l \in \mathbb{H}_l$ and is defined as
$$y^{l^\prime} = h \circ (\sum_{l\in A}O_{K_l}(y^l))$$
\end{itemize}
\end{defn}

Therefore, by Proposition 2, we have proved:
\newline
\textbf{Corollary 2} \emph{Let $z(\cdot) \sim \mathcal{SP}(0,C)$ be a centered continuous stochastic process on $\mathbb{H} = \mathcal{L}^2(\mathbb{R}^d)$. Then the Deep function machine operator $g = O_{DFM}(z,S)$ defines a well-defined stochastic process on $\mathbb{H}$. }

\subsection{Inverse Wishart process as a prior for kernel functions}\label{app:wish}
\begin{defn} [Inverse Wishart processes \cite{shah2014student}] Let $\Sigma$ be random function $\Sigma(\cdot,\cdot):\mathcal{X}\times\mathcal{X} \rightarrow \mathbb{R}$. A stochastic process defined on such functions is called the inverse Wishart process on $\mathcal{X}$ with parameter $\nu$ and base function $\Psi:\mathcal{X}\times\mathcal{X} \rightarrow \mathbb{R}$, if for any finite collection of input data $\Xvec=\{\xvec_s\}_{1\leq s \leq N_s}$, the corresponding matrix-valued evaluation $\Sigma(\Xvec,\Xvec)\in\Pi(N_s)$ is distributed according to an inverse Wishart distribution $\Sigma(\Xvec,\Xvec)\sim \text{IW}_S(\nu,\Psi(\Xvec,\Xvec))$. We denote $\Sigma\sim\mathcal{IWP}(v,\Psi(\cdot,\cdot))$.
\end{defn}
Consider the problem in Section \ref{sec:amor} of minimizing the objective
\begin{align*}
\mathcal{U}(m,\Kcal) = \mathcal{D}_{\text{KL}}[p(\fvec,\yvec|\Xvec,\theta)||q_{\mathcal{GP}}(\fvec, \yvec|\Xvec, m(\cdot),\Kcal(\cdot,\cdot))] 
\end{align*}
Since we use $q(\yvec|\fvec) = p(\yvec|\fvec)$, this reduces $\mathcal{U}(m,\Kcal)$ to $\mathrm{D}_{KL}[p(\fvec | \Xvec, \theta) || q_{\mathcal{GP}}(\fvec | \Xvec, m, \mathcal{K})]$. In order to obtain optimal solution wrt. $\mathcal{U}(m,\Kcal)$, it sufficies to draw $S$ fantasy functions (each sample is a random function $f_s(\cdot)$) from the prior distribution $p(\fvec|\Xvec,\theta)$, and perform moment matching, which gives exactly the MLE solution, i.e., empirical mean and covariance functions
\begin{align}
m_{\text{MLE}}^\star(\xvec) &= \sum_s\frac{1}{S}f_s(\xvec), \\ 
\Kcal_{\text{MLE}}^\star(\xvec_1,\xvec_2)  &= \frac{1}{S}\sum_s  \Delta_s(\xvec_1) \Delta_s(\xvec_2), \\
\Delta_s(\xvec) &= f_s(\xvec)-m_{\text{MLE}}^\star(\xvec).
\end{align}
In practice, in order to gain computational advantage, the number of fantasy functions $S$ is often small, therefore we further put an inverse wishart process prior over the GP covariance function, i.e. $\Kcal(\cdot,\cdot) \sim \mathcal{IWP}(\nu,\Psi)$. By doing so, we are able to give MAP estimation instead of MLE estimation. Since inverse Wishart distribution is conjugate to multivariate Gaussian distribution, the maximum a posteriori (MAP) solution is given by
\begin{align}
\nonumber &\Kcal_{\text{MAP}}^\star(\xvec_1,\xvec_2) \\
& = \frac{1}{\nu+S+N+1} \{ \sum_s  \Delta_s(\xvec_1) \Delta_s(\xvec_2) 
+ \Psi(\xvec_1,\xvec_2) \}.
\end{align}
Where $N$ is the number of data points in the training set $\Xvec$ where $m(\cdot)$ and $\Kcal(\cdot,\cdot)$ are evaluated. Alternatively, one could also use the posterior mean Estimator (PM) that minimizes posterior expected squared loss:
\begin{align}
\nonumber & \Kcal_{\text{PM}}^\star(\xvec_1,\xvec_2)  \\
&= \frac{1}{\nu+S-N-1} \{ \sum_s  \Delta_s(\xvec_1) \Delta_s(\xvec_2) + \Psi(\xvec_1,\xvec_2) \}.
\end{align}
In the implementation of this paper, we choose $\Kcal_{\text{PM}}$ estimator with $\nu = N$ and $\Psi(\xvec_1,\xvec_2) = \psi\delta(\xvec_1,\xvec_2)$. The hyper parameter $\psi$ is trained using fast grid search using the same procedure for the noise variance parameter, as detailed in Appendix \ref{app:imp}.

\subsection{Derivation of the upper bound \texorpdfstring{$\mathcal{U}(m,\Kcal)$} for sleep phase}
\label{app:ub}
Applying the chaine rule of KL-divregence, we have
\begin{align*}
\mathcal{J}(m,\Kcal) = &\mathcal{D}_{\text{KL}}[p(\fvec|\Xvec,\yvec,\theta)||q_{\mathcal{GP}}(\fvec|\Xvec, \yvec,m(\cdot),\Kcal(\cdot,\cdot))] \\
= &\mathcal{D}_{\text{KL}}[p(\fvec,\yvec|\Xvec,\theta)||q_{\mathcal{GP}}(\fvec,\yvec |\Xvec,m(\cdot),\Kcal(\cdot,\cdot))] \\
&-\mathcal{D}_{\text{KL}}[p(\yvec|\Xvec,\theta)||q_{\mathcal{GP}}(\yvec|\Xvec, m(\cdot),\Kcal(\cdot,\cdot))] \\ 
= & \mathcal{U}(m,\Kcal) - \mathcal{D}_{\text{KL}}[p(\yvec|\Xvec,\theta)||q_{\mathcal{GP}}(\yvec|\Xvec, m(\cdot),\Kcal(\cdot,\cdot))].
\end{align*}
Therefore, by the non-negative property of KL divergence, we have $\mathcal{J}(m,\Kcal) < \mathcal{U}(m,\Kcal)$.
Since we select $q(\yvec | \fvec) = p(\yvec | \fvec)$, the optimal solution of $\mathcal{U}(m,\Kcal)$ also minimizes $\mathcal{D}_{\text{KL}}(p(\yvec|\Xvec,\theta)||q_{\mathcal{GP}}(\yvec|\Xvec, m(\cdot),\Kcal(\cdot,\cdot)))$. Therefore not only the upper bound $\mathcal{U}$ is optimized in sleep phase, the gap $-\mathcal{D}_{\text{KL}}(p(\yvec|\Xvec,\theta)||q_{\mathcal{GP}}(\yvec|\Xvec, m(\cdot),\Kcal(\cdot,\cdot)))$ is also decreased when the mean and covariance functions are optimized. 


\subsection{Empirical Bayes approximation for VIP with a hierarchical prior on $
\theta$}
\label{app:full}
The implicit processes (such as Bayesian neural networks and GPs) could be sensitive to the choice of the model parameters (that is, parameters $\theta$ of the prior). To make our variational implicit process more robust we further present an empirical Bayesian treatment, by introducing an extra hierarchical prior distribution $p(\theta)$ on the prior parameters $\theta$, and fitting a variational approximation $q(\theta)$ to the posterior. Sleep phase updates remain the same when conditioned on a given configuration of $\theta$.
The $\alpha$-energy term in wake phase learning becomes
\begin{equation}
\begin{aligned}
& \log q_{\mathcal{GP}}(\yvec|\Xvec) \\
& =\log \int_{\theta}q_{\mathcal{GP}}(\yvec|\Xvec,\theta) p(\theta)d\theta \approx \mathcal{L}_{\mathcal{GP}}^{\alpha}
(q(\avec),q(\theta)), \\
&\mathcal{L}_{\mathcal{GP}}^{\alpha}
(q(\avec),q(\theta)) \\
&= \frac{1}{\alpha}\sum_n^N \log \mathbb{E}_{q(\avec)q(\theta)} \left[ q^\star(y_n|\xvec_n, \avec, \theta)^\alpha \right] \\
& - \mathrm{D}_{\text{KL}}[q(\avec)||p(\avec)] - \mathrm{D}_{\text{KL}}[q(\theta)||p(\theta)].
\end{aligned}
\end{equation}
Compared with the approximate MLE method, the only extra term needs to be estimated is $- \mathcal{D}_{\text{KL}}[q(\theta)||p(\theta)]$. Note that, introducing $q(\theta)$ will double the number of parameters. In the case of Bayesian NN as an IP, where $\theta$ contains means and variances for weight priors, then a simple Gaussian $q(\theta)$ will need two sets of means and variances variational parameters (i.e., posterior means of means, posterior variances of means,posterior means of variances, posterior variances of variances). Therefore, to make the representation compact, we choose $q(\theta)$ to be a Dirac-delta function $\delta(\theta_q)$, which results in an \emph{empirical Bayesian solution.}

Another possible alternative approach is, instead of explicitly specifying the form and hyperparameters for $p(\theta)$,we can notice that from standard variational lower bound
\begin{align*}
\log q_{\mathcal{GP}}(\yvec|\Xvec) \approx \mathbb{E}_{q(\theta)}[ \log q_{\mathcal{GP}}(\yvec|\Xvec,\theta) ] - \mathcal{D}_{\text{KL}}[q(\theta)||p(\theta)].
\end{align*}
Then $\mathcal{D}_{\text{KL}}[q(\theta)||p(\theta)]$ can be approximated by 
\begin{align*}
- \mathcal{D}_{\text{KL}}[q(\theta)||p(\theta)] &\approx -\mathbb{E}_{q(\theta)}[ \log q_{\mathcal{GP}}(\yvec|\Xvec,\theta) ] + \text{constant} \\
& = -\log q_{\mathcal{GP}}(\yvec|\Xvec,\theta_q)  + \text{constant}
\end{align*}
Therefore, we can use $-\log q_{\mathcal{GP}}(\yvec|\Xvec,\theta_q)$ as the regularization term instead, which penalizes the parameter configurations that returns a \emph{full} marginal log likelihood (as opposed to the diagonal likelihood in the original BB-$\alpha$ energy  $\frac{1}{\alpha}\sum_n^N \log \mathbb{E}_{q(\zvec)q(\theta)} q_{\mathcal{GP}}(y_n|\xvec_n, \zvec, \theta)^\alpha$) that is too high, especially the contribution from non-diagonal covariances. We refer this as \emph{likelihood regularization}. In practice, $-\log q_{\mathcal{GP}}(\yvec|\Xvec,\theta_q)$ is estimated on each mini-batch.

\section{KL divergence on function space v.s. KL divergence on weight space}

We briefly discuss KL divergence on function space in finite dimensional case. In the sleep phase of VIP, we have proposed minimizing the following KL divergence in function space:
\begin{align}
    \mathcal{U}(m,\Kcal) = \mathrm{D}_{\text{KL}}[p(\yvec, \fvec | \Xvec, \theta) || q_{\mathcal{GP}}(\yvec, \fvec | \Xvec, m, \mathcal{K})].
\end{align}

This is an example of KL divergence in function space (i.e., the output $\fvec$). Generally speaking, we may assume that $p(\fvec) = \int_{\mathbf{W}}p(\fvec|\mathbf{W})p(\mathbf{W})d\mathbf{W}$, and $q(\fvec)  =  \int_{\mathbf{W}}p(\fvec|\mathbf{W})q(\mathbf{W})$, where $q(\mathbf{W})$ is weight-space variational approximation. That is to say, both stochastic processes $p$ and $q$ can be generated by finite dimensional weight space representation $\mathbf{W}$. This can be seen as a one-step Markov chain with preivious state $s_t = \mathbf{W}$, new state $s_{t+1} = \fvec$, and probability transition function $r(s_{t+1}|s_t) = p(\fvec|\mathbf{W})$. Then, by applying the second law of thermodynamics of Markov chains(\citet{cover2012elements}), we have:
\begin{align}
    \mathrm{D}_{\text{KL}}[p(\fvec) || q(\fvec)] \leq \mathrm{D}_{\text{KL}}[p(\mathbf{W}) || q(\mathbf{W})]
\end{align}
This shows that the KL divergence in function space forms a tighter bound than the KL divergence on weight space, which is one of the merits of function space inference.

\section{Further discussions on Bayesian neural networks}
\label{sec:app_compare}
We provide a comparison between our kernel in equation (\ref{eq:K}), and the kernel proposed in \citet{gal2016dropout}. Notably, consider the following Gaussian process:
\begin{align}
\nonumber &y(\cdot) \sim \mathcal{GP}(0, \Kcal_{\text{VDO}}(\cdot, \cdot)), \\
\nonumber &\Kcal_{\text{VDO}}(\xvec_1,\xvec_2) = \\
&\int p(\mathbf{w})p(b)\sigma(\mathbf{w}^\top\xvec_1+b)\sigma(\mathbf{w}^\top\xvec_2+b) d\mathbf{w}db.
\label{eq:kernel_gp}
\end{align}
Here $\sigma(\cdot)$ is a non-linear activation function, $\mathbf{w}$ is a vector of length $D$, $b$ is the bias scaler, and $p(\mathbf{w})$, $p(b)$ the corresponding prior distributions. 
\citet{gal2016dropout} considered approximating this GP with a one-hidden layer BNN $\hat{y}(\cdot) = \text{BNN}(\cdot, \theta)$ with $\theta$ collecting the weights and bias vectors of the network. Denote the weight matrix of the first layer as $\bm{W} \in \mathbb{R}^{D \times K}$, i.e.~the network has $K$ hidden units, and the $k$th column of $\bm{W}$ as $\mathbf{w}_k$. Similarly the bias vector is $\bm{b} = (b_1, ..., b_K)$. We further assume the prior distributions of the first-layer parameters are $p(\bm{W}) = \prod_{k=1}^K p(\mathbf{w}_k)$ and $p(\bm{b}) = \prod_{k=1}^K p(b_k)$, and use mean-field Gaussian prior for the output layer. Then this BNN constructs an approximation to the GP kernel as:
\begin{align*}
\tilde{\Kcal}_{\text{VDO}}(\xvec_1,\xvec_2) 
&= \frac{1}{K}\sum_k \sigma(\mathbf{w}_k^\top\xvec_1+b_k)\sigma(\mathbf{w}_k^\top\xvec_2+b_k), \\ 
\mathbf{w}_k &\sim p(\mathbf{w}), \quad b_k \sim p(b). \label{eq:K_VDO}
\end{align*}
This approximation is equivalent to the empirical estimation (\ref{eq:K}), if $S=K$ and the IP is defined by 
\begin{equation}
\begin{aligned}
g_{\theta}(\xvec, \zvec) &= \sigma(\mathbf{w}^\top \xvec + b), \zvec = \{\mathbf{w}, b\}, p(\zvec) = p(\mathbf{w})p(b), \\
p(\zvec),& \ \sigma(\cdot) \text{ satisfy } \mathbb{E}_{p(\zvec)}[\sigma(\mathbf{w}^\top \xvec + b)] = 0.
\end{aligned}
\label{eq:implicit_no_hidden_layer}
\end{equation}
In such case, the output layer of that one-hidden layer BNN corresponds to the Bayesian linear regression ``layer'' in our final approximation.
However, the two methods are motivated in different ways. \citet{gal2016dropout} used this interpretation to approximate a GP with kernel (\ref{eq:kernel_gp}) using a one-hidden layer BNN, while our goal is to approximate the IP \ref{eq:implicit_no_hidden_layer} by a GP (note that the IP is defined as the output of the \emph{hidden} layer, not the output of the BNN). Also this coincidence only applies when the IP is defined by a Bayesian logistic regression model, and our approximation is applicable to BNN and beyond.


\section{Further experimental details}\label{app:imp}
We provide further experimental details in this section. We opensource the code of VIP for UCI experiments at \url{https://github.com/LaurantChao/VIP}.

\subsection{General settings for VIP}
For small datasets we use the posterior GP equations for prediction, otherwise we use the $\mathcal{O}(S^3)$ approximation. We use $S=20$ for VIP unless noted otherwise. When the VIP is equipped with a Bayesian NN/LSTM as prior over functions (Example \ref{ex:bnn}-\ref{ex:lstm}), the prior parameters over each weight are untied, thus can be individually tuned. Empirical Bayesian estimates of the prior parameters are used in \ref{exp:reg} and \ref{exp:lstm}. 

\subsection{Further experimental details of synthetic example}
The compositional kernel for GP is the summation of RBF and Periodic kernels.
In this toy experiment, both VDO and VIP use a BNN as the underlying model. Note that it appears that the GP slightly overfits. It is possible to hand-pick the kernel parameters for a smoother fit of GP. However, we have found that quantitatively this will result in a decrease in test predictive likelihood and an increase of RMSE. Therefore, we chose to optimize the kernel parameters by maximizing the marginal likelihood. 

\subsection{Further implementation details for multivariate regression experiments}
\begin{itemize}
\item Variational Gaussian inference for BNN (VI-BNN): we implement VI for BNN using the Bayesian deep learning library, ZhuSuan \cite{shi2017zhusuan}. VI-BNN employs a mean-field Gaussian variational approximation but evaluates the variational free energy using the reparameterisation trick \cite{kingma2013auto}. We use a diagonal Gaussian prior for the weights and fix the prior variance to 1. The noise variance of the Gaussian noise model is optimized together with the means and variances of the variational approximation using the variational free energy.
\item Variational implicit process-Neural Sampler regressor (VIP-NS): we use neural sampler with two hidden layers of 10 hidden units. The input noise dimension is 10 or 50, which is determined using validation set.
\item Variational dropout (VDO) for BNN: similar to \citet{gal2016dropout}, we fix the length scale parameter $0.5*l^2 = 10e^{-6}$. Since the network size is relatively small, dropout probability is set as 0.005 or 0.0005. We use 2000 forward passes to evaluate posterior likelihood.
\item $\alpha$-dropout inference for BNN: suggested by \citet{li2017dropout}, we fix $\alpha=0.5$ which often gives high quality uncertainty estimations, possibility due to it is able to achieve a balance between reducing training error and improving predictive likelihood. We use $K=10$ for MC sampling.
\item Variational sparse GPs and exact GPs: we implement the GP-related algorithms using GPflow \cite{matthews2017gpflow}. variational sparse GPs uses 50 inducing points. Both GP models use the RBF kernel.
\item About noise variance parameter grid search for VIPs (VIP-BNN and VIP-NS), VDOs and $\alpha$-dropout: we start with random noise variance parameter, run optimization on the model parameters, and then perform a (thick) grid search over noise variance parameter on validation set. Then, we train the model on the entire training set using this noise variance parameter value. This coordinate ascent like procedure does not require training the model for multiple times as in Bayesian optimization, therefore can speed up the learning process. The same procedure is used to search for optimal hyperparameter $\psi$ of the inverse-Wishart process of VIPs.
\end{itemize}

\subsection{Additional implementation details for ABC experiment}

Following the experimental setting of \citet{papamakarios2016fast}, we set the ground truth L-V model parameter to be $\theta_1 = 0.01, \theta_2 = 0.5,\theta_3 = 1.0,\theta_4 = 0.01$. We simulate population data in the range of $[0,30]$ with step size 0.05, which result in 600 gathered measurements. We use the first 500 measurements as training data, and the remaining as test set. For MCMC-ABC and SMC-ABC setup, we also follow the implementation of \citet{papamakarios2016fast}.\footnote{\url{https://github.com/gpapamak/epsilon_free_inference}}
MCMC-ABC is ran for 10000 samples with tolerance $\epsilon$ set to be 2.0 which is manually tuned to give the best performance. In MCMC-ABC, last 100 samples are taken as samples. Likewise SMC-ABC uses 100 particles. Model likelihood is calculated based on Gaussian fit. VIP ($\alpha = 0$) is trained for 10000 iterations with Adam optimizer using 0.001 learning rate.

\subsection{Additional implementation details for predicting power conversion efficiency of organic photovoltaics molecules}
For Bayesian LSTMs, we put Gaussian prior distributions over LSTM weights. The output prediction is defined as the final output at the last time step of the input sequence. We use $S=10$ for VIP. All methods use Adam with a learning rate of 0.001 for stochastic optimization. Noise variance parameter are not optimized, but set to suggested value according to \citet{hernandez2016black}.To match the run time of the fingerprint-based methods, all LSTM methods are trained for only 100 epochs with a batch size of 250. Among different models in the last few iterations of optimization, we choose the one with the best training likelihood for testing. Note that in the original paper of variational dropout and $\alpha$-dropout inference, $K$ sample paths ($K=1$ for VDO and $K=10$ for $\alpha$-dropout) are created for \emph{each} training data, which is too prohibitive for memory storage. Therefore, in our implementation, we enforce all training data to share $K$ sample paths. This approximation is accurate since we use a small dropout rate, which is 0.005.

\subsection{Additional Tables}

\begin{table}[ht]
\centering
\caption{Interpolation performance on toy dataset.}
\vspace{-0.15in}
\scalebox{0.8}{
\begin{tabular}{p{1.in}p{0.8in}p{0.8in}p{0.8in}} \toprule
\textbf{Method} &VIP & VDO & GP \\ \midrule
Test NLL & \textbf{-0.60$\pm$0.01} & $-0.07\pm0.01$ &$-0.27\pm0.00$ \\ 
Test RMSE& \textbf{0.140$\pm$0.00} &0.161$\pm$0.00 & 0.152$\pm$0.00 \\ \bottomrule
\end{tabular}
}
\label{tab:toy}
\end{table}

\begin{table}[ht]
\centering
\caption{Interpolation performance on solar irradiance.}
\vspace{-0.15in}
\scalebox{0.75}{
\begin{tabular}{p{1.in}p{0.7in}p{0.7in}p{0.7in}p{0.7in}} \toprule
\textbf{Method} &VIP & VDO & SVGP & GP \\ \midrule
Test NLL & \textbf{0.08$\pm$0.02} & $0.21\pm0.04$ & $0.56\pm 0.23$ & $0.832\pm 0.00$ \\ 
Test RMSE& \textbf{0.28$\pm$0.00} &0.29$\pm$0.01 & 0.55$\pm$0.08 & 0.650$\pm$0.0 \\ \bottomrule
\end{tabular}
}
\label{tab:solar}
\end{table}

\begin{table}[ht]
\vspace*{-10pt}
\label{tab:cep}
\centering
\caption{Performance on clean energy dataset}
\label{tab:cep}
\vspace{-0.15in}
\scalebox{0.6}{
\begin{tabular}{l@{\ica}r@{$\pm$}l@{\ica}r@{$\pm$}l@{\ica}r@{$\pm$}l@{\ica}r@{$\pm$}l@{\ica}r@{$\pm$}l@{\ica}r@{$\pm$}l@{\ica}r@{$\pm$}l@{\ica}}
\hline
\bf{Metric}&\multicolumn{2}{c}{\bf{VIP}}&\multicolumn{2}{c}{\bf{VDO-LSTM}}&\multicolumn{2}{c}{\bf{$\alpha$-LSTM}}&\multicolumn{2}{c}{\bf{BB-$\alpha$}}&\multicolumn{2}{c}{\bf{VI-BNN}}&\multicolumn{2}{c}{\bf{FITC-GP}}&\multicolumn{2}{c}{\bf{EP-DGP}}\\
\hline
Test NLL&\textbf{0.65}&\textbf{0.01}&1.24&0.01&2.06&0.02&0.74&0.01&1.37&0.02&1.25&0.00&0.98&0.00\\
Test RMSE&\textbf{0.88}&\textbf{0.02}&0.93&0.01&1.38&0.02&1.08&0.01&1.07&0.01&1.35&0.00&1.17&0.00\\\hline
\end{tabular}
}
\end{table}

\end{appendices}

\end{document}